\documentclass[runningheads]{llncs}
\usepackage{hyperref}       
\usepackage{url}            
\usepackage{booktabs}       
\usepackage{amsfonts}       
\usepackage{amsmath}
\usepackage{nicefrac}       
\usepackage{microtype}      
\usepackage{bm}
\usepackage{caption}
\usepackage{subcaption}
\usepackage{graphicx}
\usepackage{multirow}
\usepackage{wrapfig}
\usepackage{xcolor}
\usepackage{fontawesome}
\usepackage{mathtools}
\usepackage[linesnumbered,ruled]{algorithm2e}
\usepackage[utf8x]{inputenc} 

\newcommand\defeq{\stackrel{\mathclap{\normalfont\mbox{def}}}{=}}

\begin{document}
\title{Joslim: \underline{J}oint Widths and Weights \underline{O}ptimization for \underline{Slim}mable Neural Networks}
%
%
\author{Ting-Wu Chin\inst{1}$^{(\text{\faEnvelopeO})}$ \and
Ari S. Morcos\inst{2} \and
Diana Marculescu\inst{1,3}}
\authorrunning{T.-W. Chin et al.}
%
\institute{Department of ECE, Carnegie Mellon University, Pittsburgh PA, USA
\and
Facebook AI Research, Menlo Park CA, USA
\and
Department of ECE, The University of Texas at Austin, Austin TX, USA\\
\email{\faEnvelopeO~tingwuc@alumni.cmu.edu}}

\maketitle              
\begin{abstract}
    Slimmable neural networks provide a flexible trade-off front between prediction error and computational requirement (such as the number of floating-point operations or FLOPs) with the same storage requirement as a single model. They are useful for reducing maintenance overhead for deploying models to devices with different memory constraints and are useful for optimizing the efficiency of a system with many CNNs. However, existing slimmable network approaches either do not optimize layer-wise widths or optimize the shared-weights and layer-wise widths independently, thereby leaving significant room for improvement by joint width and weight optimization. In this work, we propose a general framework to enable joint optimization for both width configurations and weights of slimmable networks. Our framework subsumes \textit{conventional and NAS-based slimmable methods} as special cases and provides flexibility to improve over existing methods. From a practical standpoint, we propose Joslim, an algorithm that jointly optimizes both the widths and weights for slimmable nets, which outperforms existing methods for optimizing slimmable networks across various networks, datasets, and objectives. Quantitatively, improvements up to 1.7\% and 8\% in top-1 accuracy on the ImageNet dataset can be attained for MobileNetV2 considering FLOPs and memory footprint, respectively. Our results highlight the potential of optimizing the channel counts for different layers \emph{jointly} with the weights for slimmable networks. Code available at \url{https://github.com/cmu-enyac/Joslim}.
\end{abstract}
\section{Introduction}
Slimmable neural networks have been proposed with the promise of enabling multiple neural networks with different trade-offs between prediction error and the number of floating-point operations (FLOPs), \textit{all at the storage requirement of only a single neural network}~\cite{yu2018slimmable}. This is in stark contrast to channel pruning methods~\cite{berman2020aows,yu2019autoslim,guo2020dmcp} that aim for a small standalone model. Slimmable neural networks are useful for applications running on mobile and other resource-constrained devices. As an example, the ability to deploy multiple versions of the same neural network alleviates the maintenance overhead for applications which support a number of different mobile devices with different memory and storage constraints, as only one model needs to be maintained. On the other hand, slimmable networks can bee critical for designing an efficient system that runs multiple CNNs. Specifically, an autonomous robot may execute multiple CNNs for various tasks at the same time. When optimizing the robot's efficiency (overall performance \textit{vs.} computational costs), it is unclear which CNNs should be trimmed by how much to achieve an overall best efficiency. As a result, methods based on trial-and-error are necessary for optimizing such a system. However, if trimming the computational requirement of any CNN requires re-training or fine-tuning, this entire process will be impractically expensive. In this particular case, if we replace each of the CNNs with their respective slimmable versions, optimizing a system of CNNs becomes practically feasible as slimmable networks can be slimmed without the need of re-training or fine-tuning.

A slimmable neural network is trained by simultaneously considering networks with different widths (or filter counts) using a single set of shared weights. The width of a child network is specified by a real number between 0 and 1, which is known as the ``width-multiplier''~\cite{howard2017mobilenets}. Such a parameter specifies how many filters per layer to use proportional to the full network. For example, a width-multiplier of $0.35\times$ represents a network with 35\% of the channel counts of the full network for all the layers. While specifying child networks using a single width-multiplier for all the layers has shown empirical success~\cite{yu2019universally,yu2018slimmable}, such a specification neglects that different layers affect the network's output differently~\cite{zhang2019all} and have different FLOPs and memory footprint requirements~\cite{gordon2018morphnet}, which may lead to sub-optimal results. As an alternative, neural architecture search (NAS) methods such as BigNAS~\cite{yu2020bignas} optimizes the layer-wise widths for slimmable networks, however, a sequential greedy procedure is adopted to optimize the widths and weights. As a result, the optimization of weights is not adapted to the optimization of widths, thereby leaving rooms for improvement by joint width and weight optimization.

\begin{figure*}[t]
    \centering
    \includegraphics[width=1\textwidth]{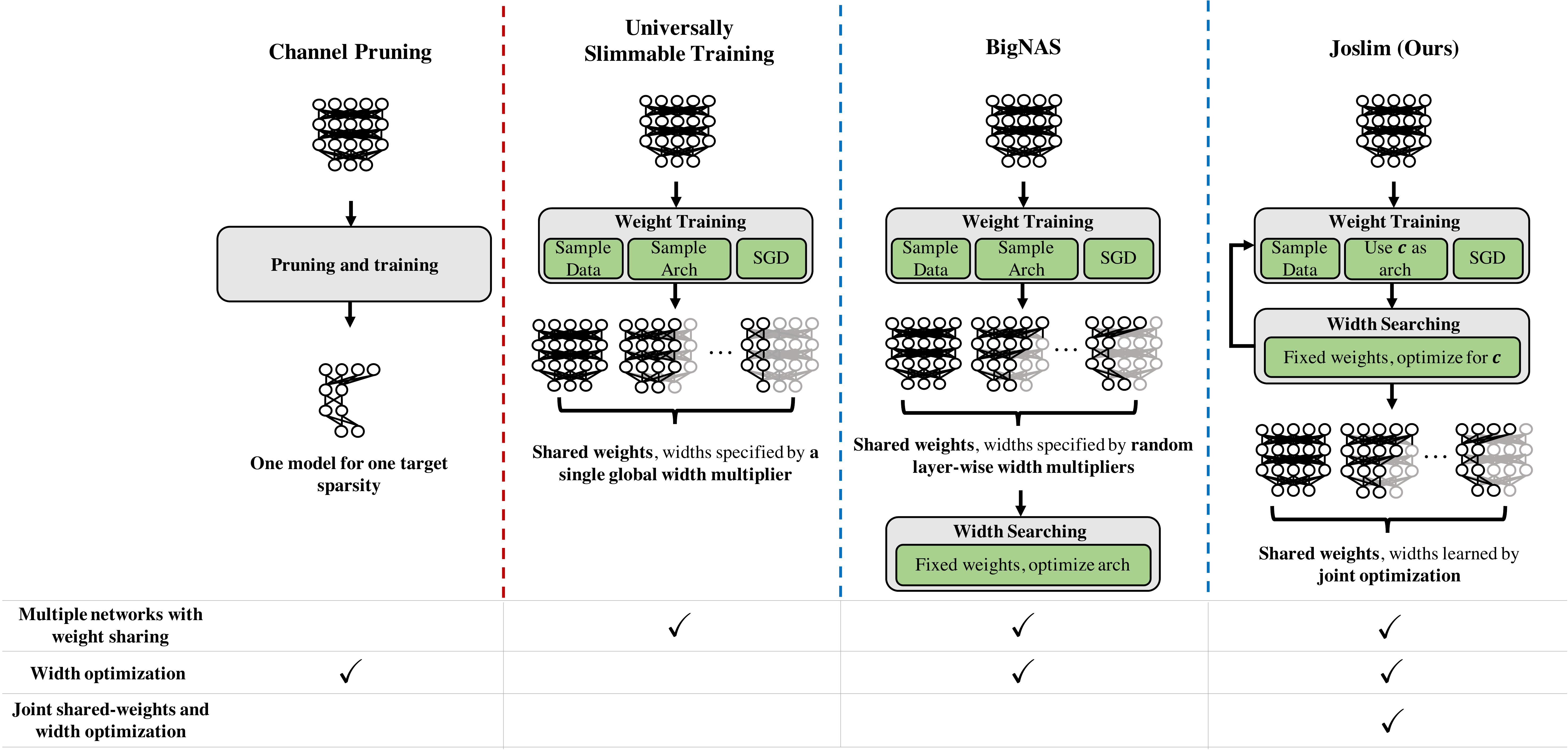}
    \caption{Schematic overview comparing our proposed method with existing alternatives and channel pruning. Channel pruning has a fundamentally different goal compared to ours, \textit{i.e.}, training slimmable nets. Joslim jointly optimizes both the widths and the shared weights.}
    \label{fig:scheme}
\end{figure*}

In this work, we propose a framework for optimizing slimmable nets by formalizing it as minimizing the area under the trade-off curve between prediction error and some metric of interest, \textit{e.g.}, memory footprint or FLOPs, with alternating minimization. Our framework subsumes both the universally slimmable networks~\cite{yu2019universally} and BigNAS~\cite{yu2020bignas} as special cases. The framework is general and provides us with insights to improve upon existing alternatives and justifies our new algorithm Joslim, the first approach that jointly optimizes both shared-weights and widths for slimmable nets. To this end, we demonstrate empirically the superiority of the proposed algorithm over existing methods using various datasets, networks, and objectives. We visualize the algorithmic differences between the proposed method and existing alternatives in Fig.~\ref{fig:scheme}.

The contributions of this work are as follows:
\begin{itemize}
    \item We propose a general framework that enables the joint optimization of the widths and their corresponding shared weights of a \textit{slimmable net}. The framework is general and subsumes existing algorithms as special cases.
    \item We propose Joslim, an algorithm that jointly optimizes the widths and weights of slimmable nets. We show empirically that Joslim outperforms existing methods on various networks, datasets, and objectives. Quantitatively, improvements up to 1.7\% and 8\% in top-1 accuracy on ImageNet are attained for MobileNetV2 considering FLOPs and memory footprint, respectively.
\end{itemize}

\section{Related work}
\subsection{Slimmable neural networks}
Slimmable neural networks~\cite{yu2018slimmable} enable multiple sub-networks with different compression ratios to be generated from a single network with one set of weights. This allows the network FLOPs to be dynamically configurable at run-time without increasing the storage requirement of the model weights. Based on this concept, better training methodologies have been proposed to enhance the performance of slimmable networks~\cite{yu2019universally}. One can view a slimmable network as a dynamic computation graph where the graph can be constructed dynamically with different accuracy and FLOPs profiles. With this perspective, one can go beyond changing just the width of the network. For example, one can alter the network's sub-graphs~\cite{ruiz2019adaptative}, network's depth~\cite{bolukbasi2017adaptive,huang2017multi,li2019improved,icml19shallowdeepnetworks}, and network's kernel sizes and input resolutions~\cite{Cai2020Once-for-All:,yu2020bignas,wang2021alphanet,yang2020mutualnet}. Complementing prior work primarily focusing on generalizing slimmable networks to additional architectural paradigms, our work provides the first principled formulation for jointly optimizing the weights and widths of slimmable networks. While our analysis focuses on the network widths, our proposed framework can be easily extended to other architectural parameters.

\subsection{Neural architecture search}
A slimmable neural network can be viewed as an instantiation of weight-sharing. In the literature for neural architecture search (NAS), weight-sharing is commonly adopted to reduce the search overhead~\cite{liu2018darts,stamoulis2019single,guo2019single,pmlr-v80-bender18a,berman2020aows,yu2019autoslim}. Specifically, NAS methods use weight-sharing as a proxy for evaluating the performance of the sub-networks to reduce the computational requirement of iterative training and evaluation. However, the goal of NAS is the resulting architecture as opposed to both shared-weights and architecture. Exceptions are BigNAS~\cite{yu2020bignas} and Once-for-all (OFA) networks~\cite{Cai2020Once-for-All:}; however, in neither case the architecture and shared-weights are jointly optimized. Specifically, both BigNAS and OFA employ a two-stage paradigm where the shared-weights are optimized before the architectures are optimized. This makes the trained weights oblivious to the optimized architectures.

While slimmable networks are inherently multi-objective, multi-objective optimization has also been adopted in NAS literature~\cite{dong2018dpp,cheng2018searching,lu2019nsga,elsken2018efficient,Yang_2020_CVPR}. However, a crucial difference of the present work compared to these papers is that we are interested in learning a single set of weights from which multiple FLOP configurations can be used (as in slimmable networks) rather than finding architectures independently for each FLOP configuration that can be trained from scratch freely. Put another way, in our setting, both shared-weights and the searched architecture are optimized jointly, whereas in prior work, only searched architectures were optimized.

When it comes to joint neural architecture search and weight training, ENAS~\cite{pham2018efficient} and TuNAS~\cite{bender2020can} can both be seen as joint optimization. However, in stark contrast to our work, their search is dedicated to a single network of a single computational requirement (e.g., FLOPs) while our method is designed to obtain the weights that work for various architectures across a wide range of computational requirements.

\subsection{Channel pruning}
Reducing the channel or filter counts for a pre-trained model is also known as channel pruning. In channel pruning, the goal is to find a single small model that maximizes the accuracy while satisfying some resource constraints by optimizing the layer-wise channel counts~\cite{li2016pruning,he2019filter,liu2017learning,ye2018rethinking,liu2017learning,louizos2017learning,yu2019autoslim,berman2020aows,ma2019bayesian,chin2020legr,liu2019metapruning,chin2021width}. While channel pruning also optimizes for non-uniform widths, the goal of channel pruning is crucially different from ours. The key difference is that channel pruning is concerned with a single pruned model while slimmable neural networks require a set of models to be trained using weight sharing. Nonetheless, we compare our work with pruning methods that conduct greedy channel pruning since they naturally produce in a sequence of models that have different FLOPs. In particular, we compare our work with AutoSlim~\cite{yu2019autoslim} in Appendix~\ref{app:autoslim} and demonstrate the effectiveness of our proposed Joslim.

\section{Methodology}
In this work, we are interested in jointly optimizing the network widths and network weights. Ultimately, when evaluating the performance of a slimmable neural network, we care about the trade-off curve between multiple objectives, \textit{e.g.}, theoretical speedup and accuracy. This trade-off curve is formed by evaluating the two objectives at multiple width configurations using the same shared-weights. Viewed from this perspective, both the widths and shared-weights should be optimized in such a way that the resulting networks have a better trade-off curve (\textit{i.e.}, larger area under curve). This section formalizes this idea and provides an algorithm to solve it in an approximate fashion.

\subsection{Problem formulation}
Our goal is to find both the weights and the width configurations that optimize the area under the trade-off curve between two competing objectives, \textit{e.g.,} accuracy and inference speed. Without loss of generality, we use cross entropy loss as the accuracy objective and FLOPs as the inference speed objective throughout the text for clearer context. Note that FLOPs can also be replaced by other metrics of interest such as memory footprint. Since in this case both objectives are better when lower, the objective for the optimizing slimmable nets becomes to \textit{minimize} the area under curve. To quantify the area under curve, one can use a Riemann integral. Let $w(c)$ be a width configuration of $c$ FLOPs, one can quantify the Riemann integral by evaluating the cross entropy loss $L_{\mathcal{S}}$ on the training set $\mathcal{S}$ using the shared weights $\bm{\theta}$ for the architectures that spread uniformly on the FLOPs-axis between a lower bound $l$ and an upper bound $u$ of FLOPs: $\{\bm{a} | \bm{a}=w(c), c\in[l,u]\}$. More formally, the area under curve $\mathbb{A}$ is characterized as
\begin{align}
    \mathbb{A}(\bm{\theta}, w) &\defeq \int_l^u L_{\mathcal{S}}\left(\bm{\theta}, w(c)\right) dc\\
    &\approx \sum_{i=0}^N L_{\mathcal{S}}\left(\bm{\theta}, w(c_i)\right) \delta \label{eq:2},
\end{align}
where equation~\ref{eq:2} approximates the Riemann integral with the Riemann sum using $N$ architectures that are spread uniformly on the FLOPs-axis with a step size $\delta$. With a quantifiable area under curve, our goal for optimizing slimmable neural networks becomes finding both the shared-weights $\bm{\theta}$ and the architecture function $w$ to minimize their induced area under curve:
\begin{align}
    \arg\min_{\bm{\theta}, w} \mathbb{A}(\bm{\theta},w) &\approx \arg\min_{\bm{\theta}, w} \sum_{i=0}^N L_{\mathcal{S}}\left(\bm{\theta}, w(c_i)\right) \delta\\
    &=\arg\min_{\bm{\theta}, w} \frac{1}{N} \sum_{i=0}^N L_{\mathcal{S}}\left(\bm{\theta}, w(c_i)\right) \label{eq:4}\\
    &\approx\arg\min_{\bm{\theta}, w} \mathbb{E}_{c\sim U(l,u)}  L_{\mathcal{S}}\left(\bm{\theta}, w(c)\right)\label{eq:prob},
\end{align}
where $U(l,u)$ denotes a uniform distribution over a lower bound $l$ and an upper bound $u$. Note that the solution to equation~\ref{eq:prob} is the shared-weight vector and a set of architectures, which is drastically different from the solution to the formulation used in the NAS literature~\cite{liu2018darts,tan2019mnasnet}, which is an architecture.

\subsection{Proposed approach: Joslim}\label{sec:joslim}
Since both the shared-weights $\bm{\theta}$ and the architecture function $w$ are optimization variables of two natural groups, we start by using alternating minimization:
\begin{align}
    w^{(t+1)} &= \arg\min_{w} \mathbb{E}_{c\sim U(l,u)} L_{\mathcal{S}}\left(\bm{\theta^{(t)}}, w(c)\right)\label{eq:arch}\\
    \bm{\theta^{(t+1)}} &= \arg\min_{\bm{\theta}} \mathbb{E}_{c\sim U(l,u)} L_{\mathcal{S}}\left(\bm{\theta}, w^{(t+1)}(c)\right)\label{eq:weight}.
\end{align}
In equation~\ref{eq:arch}, we maintain the shared-weights $\bm{\theta}$ fixed and for each FLOPs between $l$ and $u$, we search for a corresponding architecture that minimizes the cross entropy loss. This step can be seen as a multi-objective neural architecture search given a fixed set of pre-trained weights, and can be approximated using smart algorithms such as multi-objective Bayesian optimization~\cite{PariaKP19} or evolutionary algorithms~\cite{deb2002fast}. However, even with smart algorithms, such a procedure can be impractical for every iteration of the alternating minimization.

In equation~\ref{eq:weight}, one can use stochastic gradient descent by sampling from a set of architectures that spread uniformly across FLOPs obtained from solving equation~\ref{eq:arch}. However, training such a weight-sharing network is practically $4\times$ the training time of the largest standalone subnetwork \cite{yu2019universally} (it takes 6.5 GPU-days to train a slimmable ResNet18), which prevents it from being adopted in the alternating minimization framework.

To cope with these challenges, we propose targeted sampling, local approximation, and temporal sharing to approximate both equations.

\subsubsection{Targeted sampling} We propose to sample a set of FLOPs to approximate the expectation in equations~\ref{eq:arch} and~\ref{eq:weight} with empirical estimates. Moreover, the sampled FLOPs are shared across both steps in the alternating minimization so that one does not have to solve for the architecture function $w$ (needed for the second step), but only solve for a set of architectures that have the corresponding FLOPs. Specifically, we approximate the expectation in both equations~\ref{eq:arch} and~\ref{eq:weight} with the sample mean:
\begin{align}
    c_i^{(t)} &\sim U(l,u)~~\forall~i=1,\dots,M\\
    w^{(t+1)} &\approx \arg\min_{w} \frac{1}{M}\sum_{i=1}^M L_{\mathcal{S}}\left(\bm{\theta^{(t)}}, w(c_i^{(t)})\right)\label{eq:arch_sample}\\
    \bm{\theta^{(t+1)}} &\approx \arg\min_{\bm{\theta}} \frac{1}{M}\sum_{i=1}^M L_{\mathcal{S}}\left(\bm{\theta}, w^{(t+1)}(c_i^{(t)})\right)\label{eq:weight_sample}.
\end{align}
From equation~\ref{eq:arch_sample} and~\ref{eq:weight_sample}, we can observe that at any timestamp $t$, we only query the architecture function $w^{(t)}$ and $w^{(t+1)}$ at a fixed set of locations $c_i~\forall~i=1,\dots,M$. As a result, instead of solving for the architecture function $w$, we solve for a fixed set of architectures $\mathcal{W}^{(t+1)}$ at each timestamp as follows:
\begin{align}
    \begin{split}
        \mathcal{W}^{(t+1)} &:= \{w^{(t+1)}(c_i),\dots,w^{(t+1)}(c_M)\}
    \end{split}
\end{align}
where
\begin{align}
    \begin{split}
        w^{(t+1)}(c_i) &= \arg\min_{\bm{a}} L_{\mathcal{S}}\left(\bm{\theta^{(t)}}, \bm{a}\right)\\
        &\text{s.t.}~\text{FLOPs}(\bm{a})=c_i.
    \end{split}\label{eq:arch_set}
\end{align}
With these approximations, for each iteration in the alternating minimization, we solve for $M$ architectures with targeted FLOPs as opposed to solving for the entire approximate trade-off curve.

\subsubsection{Local approximation} To reduce the overhead for solving equation~\ref{eq:weight_sample}, we propose to approximate it with a few steps of gradient descent. Specifically, instead of training a slimmable neural network with sampled architectures until convergence in each iteration of alternating minimization (equation~\ref{eq:weight_sample}), we propose to perform $K$ steps of gradient descent: 
\begin{align}
    \begin{split}
        \bm{x^0} &\defeq \bm{\theta^{(t)}}\\
        \bm{x^{(k+1)}} &\defeq \bm{x^{(k)}} - \eta \frac{1}{M} \sum_{i=1}^M \nabla_{\bm{x^{(k)}}} L_{\mathcal{S}}\left(\bm{x^{(k)}}, \mathcal{W}^{(t+1)}_i\right)\\
        \bm{\theta^{(t+1)}} &\approx \bm{x^{(K)}},
    \end{split}\label{eq:weight_grad}
\end{align}
where $\eta$ is the learning rate. Larger $K$ indicates better approximation with higher training overhead.

\subsubsection{Temporal sharing} Since we use local approximation, $\bm{\theta^{(t+1)}}$ and $\bm{\theta^{(t)}}$ would not be drastically different. As a result, instead of performing constrained neural architecture search \textit{from scratch} (\textit{i.e.}, solving for equation~\ref{eq:arch_set}) in every iteration of the alternating minimization, we propose to share information across the search procedures in different iterations of the alternation.

To this end, we propose to perform temporal sharing for multi-objective Bayesian optimization with random scalarization (MOBO-RS)~\cite{PariaKP19} to solve equation~\ref{eq:arch_set}. MOBO-RS itself is a sequential model-based optimization algorithm, where one takes a set of architectures $\mathcal{H}$, builds models (typically Gaussian Processes~\cite{rasmussen2003gaussian}) to learn a mapping from architectures to cross entropy loss $g_{\text{CE}}$ and FLOPs $g_{\text{FLOPs}}$, scalarizes both models into a single objective with a random weighting $\lambda$ ($\lambda$ controls the preference for cross entropy and FLOPs), and finally optimizes the scalarized model to obtain a new architecture and stores the architecture back to the set $\mathcal{H}$. This entire procedure repeats for $T$ iterations for one MOBO-RS.

To exploit temporal similarity, we propose MOBO-TS2, which stands for \underline{m}ulti-\underline{o}bjective \underline{B}ayesian \underline{o}ptimization with \underline{t}argeted \underline{s}calarization and \underline{t}emporal \underline{s}haring. Specifically, we propose to let $T=1$ and share $\mathcal{H}$ across alternating minimization. Additionally, we modify the random scalarization with targeted scalarization where we use binary search to search for the $\lambda$ that results in the desired FLOPs. As such, $\mathcal{H}$ grows linearly with the number of alternations. In such an approximation, for each MOBO in the alternating optimization, we reevaluate the cross-entropy loss for each $\bm{a}\in\mathcal{H}$ to build faithful GPs. We further provide theoretical analysis for approximation via temporal similarity for Bayesian optimization in Appendix~\ref{app:bonud}.

\SetAlFnt{\small}

\begin{algorithm}
    \SetKwInOut{Input}{Input}
    \SetKwInOut{Output}{Output}
    \Input{Model parameters $\bm{\theta}$, lower bound for width-multipliers $w_0\in[0,1]$, number of full iterations $F$, number of gradient descent updates $K$, number of $\lambda$ samples $M$}
    \Output{Trained parameter $\bm{\theta}$, approximate Pareto front $\mathcal{N}$}
    $\mathcal{H} = \{\}$~~~~~~~(\textit{Historical minimizers $\bm{a}$})\\
    \For{i = 1...$F$}{
        $\bm{x}, y$ = sample\_data()\\
        $\bm{u_{\text{CE}}}$, $\bm{u_{\text{FLOPs}}}$ = $L_{\text{CE}}(\mathcal{H};\bm{\theta},\bm{x},y)$, $\text{FLOPs}(\mathcal{H})$\\
        $\bm{g_{\text{CE}}}$, $\bm{g_{\text{FLOPs}}}$ = GP\_UCB( $\mathcal{H}$, $\bm{u_{\text{CE}}}$, $\bm{u_{\text{FLOPs}}}$ )\\
        widths = []\\
        \For{m = 1...$M$}{
            $\bm{a}$ = MOBO\_TS2( $\bm{g_{\text{CE}}},\bm{g_{\text{FLOPs}}},\mathcal{H}$ )~~~~~(\textit{Algorithm~\ref{alg:pas}})\\
            widths.append($\bm{a}$)
        }
        $\mathcal{H} = \mathcal{H}~\cup$ widths~~~~~~~~~(\textit{update historical data})\\
        widths.append($\bm{w_0}$)\\
        \For{j = 1...$K$}{
            SlimmableTraining( $\bm{\theta}$, widths )\\(\textit{line 3-16 of Algorithm 1 in~\cite{yu2019universally}})
        }
        $\mathcal{N}$=nonDominatedSort($\mathcal{H}$, $\bm{u_{\text{CE}}}$, $\bm{u_{\text{FLOPs}}}$)
    }
    \caption{Joslim}\label{alg:Joslim}
\end{algorithm}

\begin{algorithm}
    \SetKwInOut{Input}{Input}
    \SetKwInOut{Output}{Output}
    \Input{Acquisition functions $\bm{g_{\text{CE}}}, \bm{g_{\text{FLOPs}}}$, historical data $\mathcal{H}$, search precision $\epsilon$}
    \Output{channel configurations $\bm{a}$}
    c = Uniform( $l, u$ )~~~~~~~(\textit{Sample a target FLOPs})\\
    $\lambda_{\text{FLOPs}}$, $\lambda_{\min}$, $\lambda_{\max}$ = 0.5, 0, 1\\
    \While(\tcp*[f]{binary search}){$| \frac{\text{FLOPs}(\bm{a}) - c}{\text{FullModelFLOPs}} | > \epsilon$}{
        $\bm{c}$=$\arg\min_c$ Scalarize( $\lambda_{\text{FLOPs}}$, $\bm{g_{\text{CE}}}, \bm{g_{\text{FLOPs}}}$ )\\
        \eIf{$\text{FLOPs}(\bm{a}) > c$}{
            $\lambda_{\min} = \lambda_{\text{FLOPs}}$\\
            $\lambda_{\text{FLOPs}} = (\lambda_{\text{FLOPs}}+\lambda_{\max})/2$\\
        }{
            $\lambda_{\max} = \lambda_{\text{FLOPs}}$\\
            $\lambda_{\text{FLOPs}} = (\lambda_{\text{FLOPs}}+\lambda_{\min})/2$\\
        }
    }
    \caption{MOBO-TS2}\label{alg:pas}
\end{algorithm}

\subsubsection{Joslim} Based on this preamble, we present our algorithm, Joslim, in Algorithm~\ref{alg:Joslim}. In short, Joslim has three steps: (1) build surrogate functions (\textit{i.e.}, GPs) and acquisition functions (\textit{i.e.}, UCBs) using historical data $\mathcal{H}$ and their function responses, (2) sample $M$ target FLOPs and solve for the corresponding widths (\textit{i.e.}, $\bm{a}$) via binary search with the scalarized acquisition function and store them back to $\mathcal{H}$, and (3) perform $K$ gradient descent steps using the solved widths. The first two steps solve equation~\ref{eq:arch_set} with targeted sampling and temporal sharing, and the final step solves equation~\ref{eq:weight_sample} approximately with local approximation. In the end, to obtain the best widths, we use non-dominated sorting based on the training loss and FLOPs for $a\in\mathcal{H}$.

\subsection{Relation to existing approaches}
For direct comparison with our work we consider the universally slimmable neural networks~\cite{yu2019universally}, which uses a single width multiplier to specify the widths of a slimmable network and NAS-based approaches such as OFA~\cite{Cai2020Once-for-All:} and BigNAS~\cite{yu2020bignas}, which have decoupled widths and weights optimization. To demonstrate the generality of the proposed framework, we show how these previously published works are special cases of our framework.
\subsubsection{Slim} Universally slimmable networks~\cite{yu2019universally}, or Slim for short, is a special case of our framework where the widths are not optimized but pre-specified by a single global width multiplier. This corresponds to solving equation~\ref{eq:prob} with $w$ given as a function that returns the width that satisfies some FLOPs by controlling a single global width multiplier. Our framework is more general as it introduces the freedom for optimizing the widths of slimmable nets.
\subsubsection{OFA and BigNAS} OFA and BigNAS use the same approach when it comes to the channel search space\footnote{Since we only search for channel counts, the progressive shrinking strategy proposed in OFA does not apply. As a result, both OFA and BigNAS have the same approach.}. They are also a special case of our framework where the optimization of the widths and the shared-weights are carried out greedily. Specifically, BigNAS first trains the shared-weights by random layer-wise width multipliers. After convergence, BigNAS performs evolutionary search to optimize the layer-wise width multipliers considering both error and FLOPs. This greedy algorithm can be seen as performing one iteration of alternating minimization by solving equation~\ref{eq:weight} followed by solving equation~\ref{eq:arch}. From this perspective, one can observe that the shared-weights $\bm{\theta}$ are not jointly optimized with the widths. Our framework is more general and enables joint optimization for both widths and weights.

As we demonstrate in Section~\ref{sec:exp}, our comprehensive empirical analysis reveals that Joslim is superior to either approach when compared across multiple networks, datasets, and objectives.

\section{Experiments}
\subsection{Experimental setup}
For all the Joslim experiments in this sub-section, we set $K$ such that Joslim only visits 1000 width configurations throughout the entire training ($|\mathcal{H}|=1000$). Also, we set $M$ to be 2, which follows the conventional slimmable training method~\cite{yu2019universally} that samples two width configurations in between the largest and the smallest widths. As for binary search, we conduct at most 10 binary searches with $\epsilon$ set to 0.02, which means that the binary search terminates if the FLOPs difference is within a two percent margin relative to the full model FLOPs. On average, the procedure terminates by using 3.4 binary searches for results on ImageNet. The dimension of $\bm{a}$ is network-dependent and is specified in Appendix~\ref{app:wm} and the training hyperparameters are detailed in Appendix~\ref{app:train}. To arrive at the final set of architectures for Joslim, we use non-dominated sort based on the training loss and FLOPs for $\bm{a}\in\mathcal{H}$.

\subsection{Performance gains introduced by Joslim}\label{sec:exp}

\begin{figure*}[t!]
     \centering
     \begin{subfigure}[b]{0.23\textwidth}
         \centering
         \includegraphics[width=\textwidth]{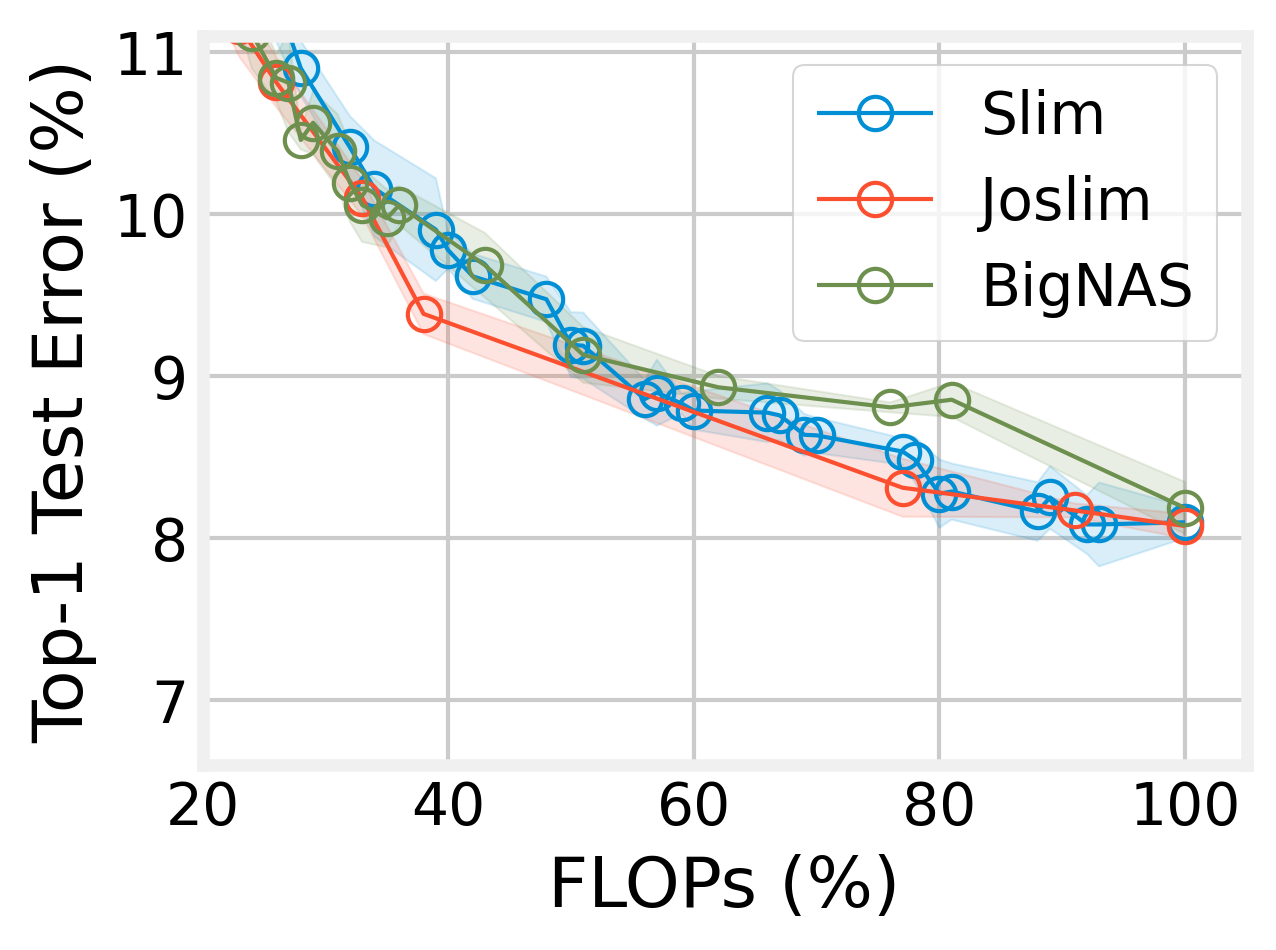}
         \caption{ResNet20 C10}
         \label{resnet20-cifar10}
     \end{subfigure}
     \hfill
     \begin{subfigure}[b]{0.23\textwidth}
         \centering
         \includegraphics[width=\textwidth]{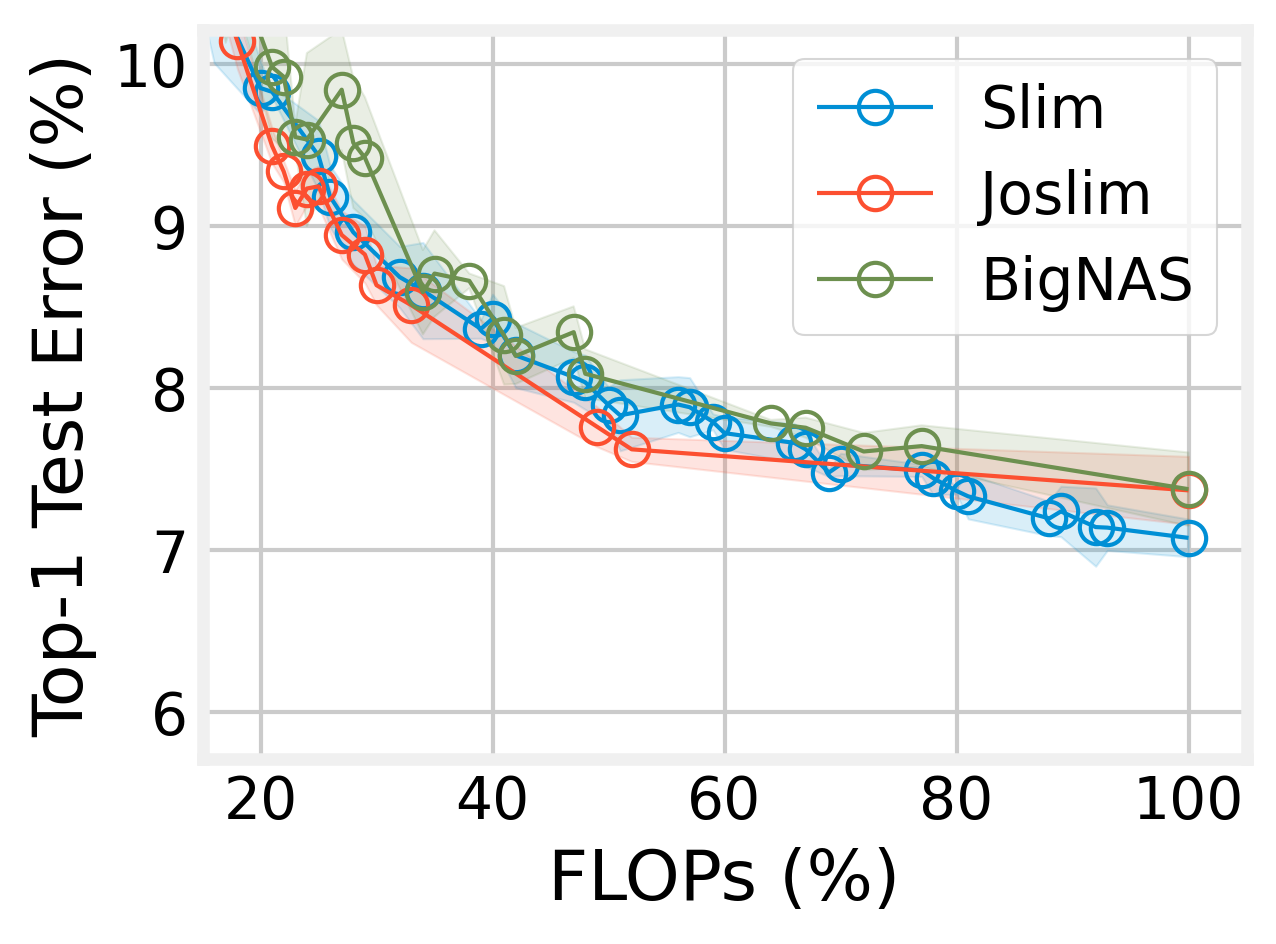}
         \caption{ResNet32 C10}
         \label{resnet32-cifar10}
     \end{subfigure}
     \hfill
     \begin{subfigure}[b]{0.23\textwidth}
         \centering
         \includegraphics[width=\textwidth]{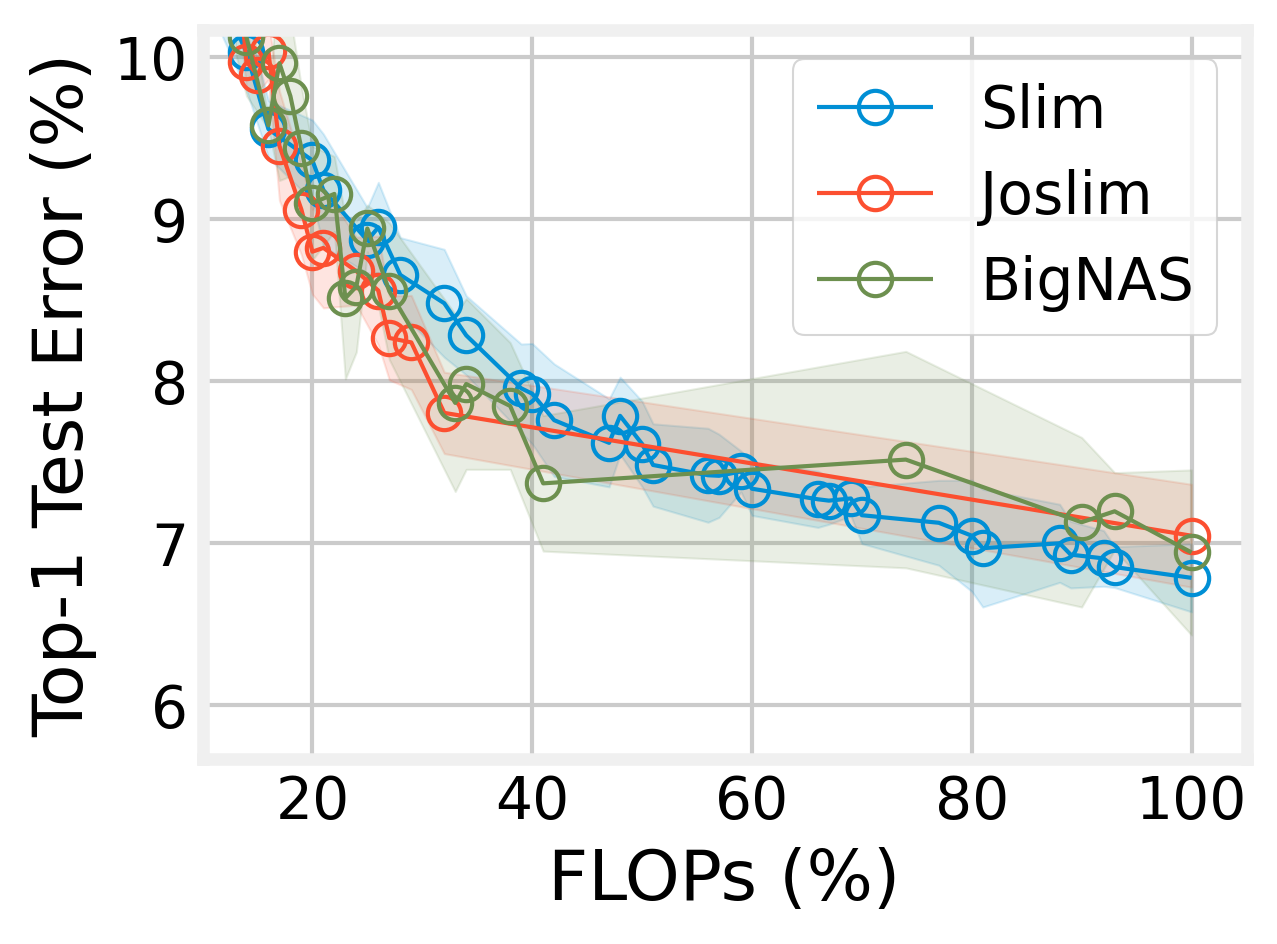}
         \caption{ResNet44 C10}
         \label{resnet44-cifar10}
     \end{subfigure}
     \hfill
     \begin{subfigure}[b]{0.23\textwidth}
         \centering
         \includegraphics[width=\textwidth]{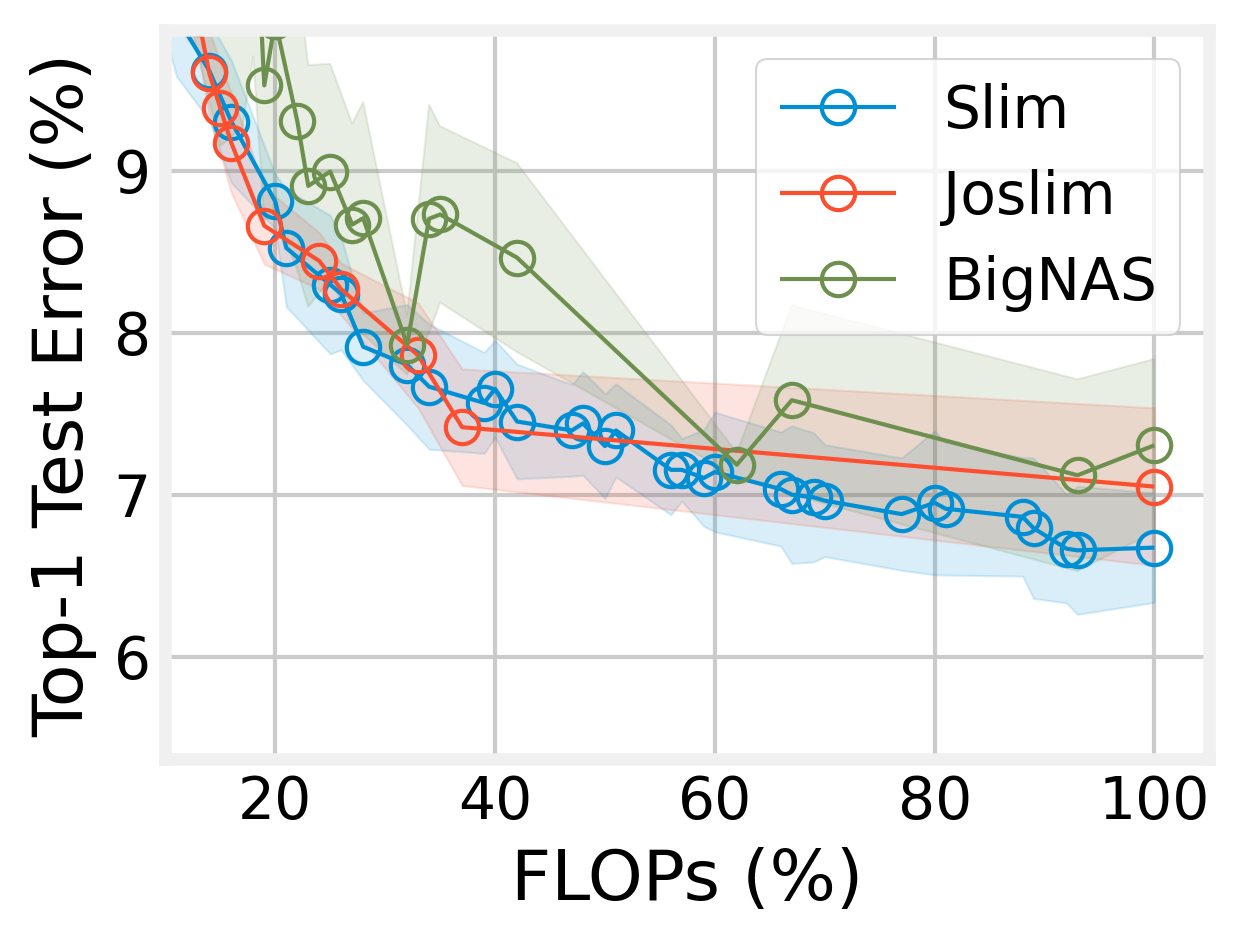}
         \caption{ResNet56 C10}
         \label{resnet56-cifar10}
     \end{subfigure}
     \\
     \begin{subfigure}[b]{0.23\textwidth}
         \centering
         \includegraphics[width=\textwidth]{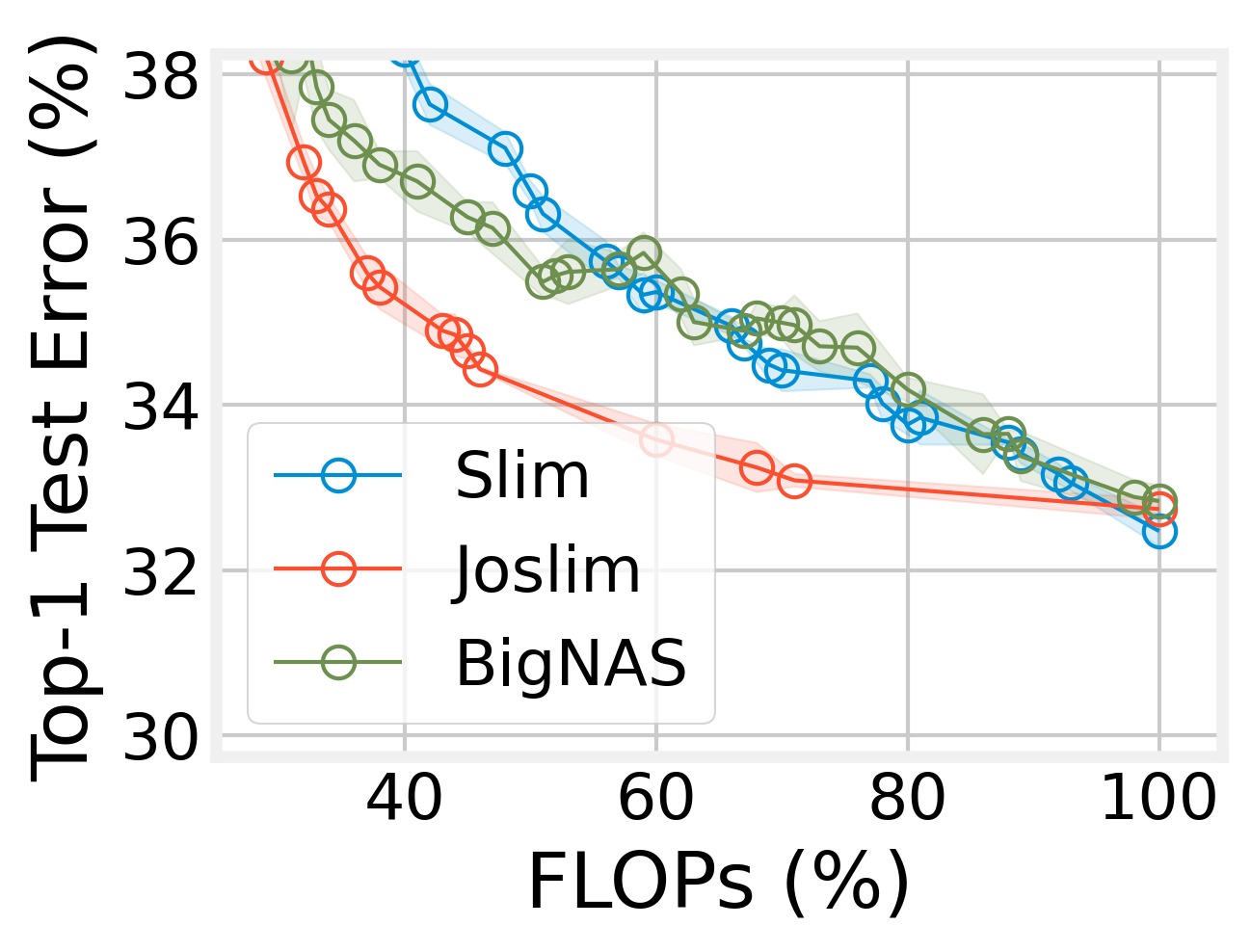}
         \caption{ResNet20 C100}
         \label{resnet20-cifar100}
     \end{subfigure}
     \hfill
     \begin{subfigure}[b]{0.23\textwidth}
         \centering
         \includegraphics[width=\textwidth]{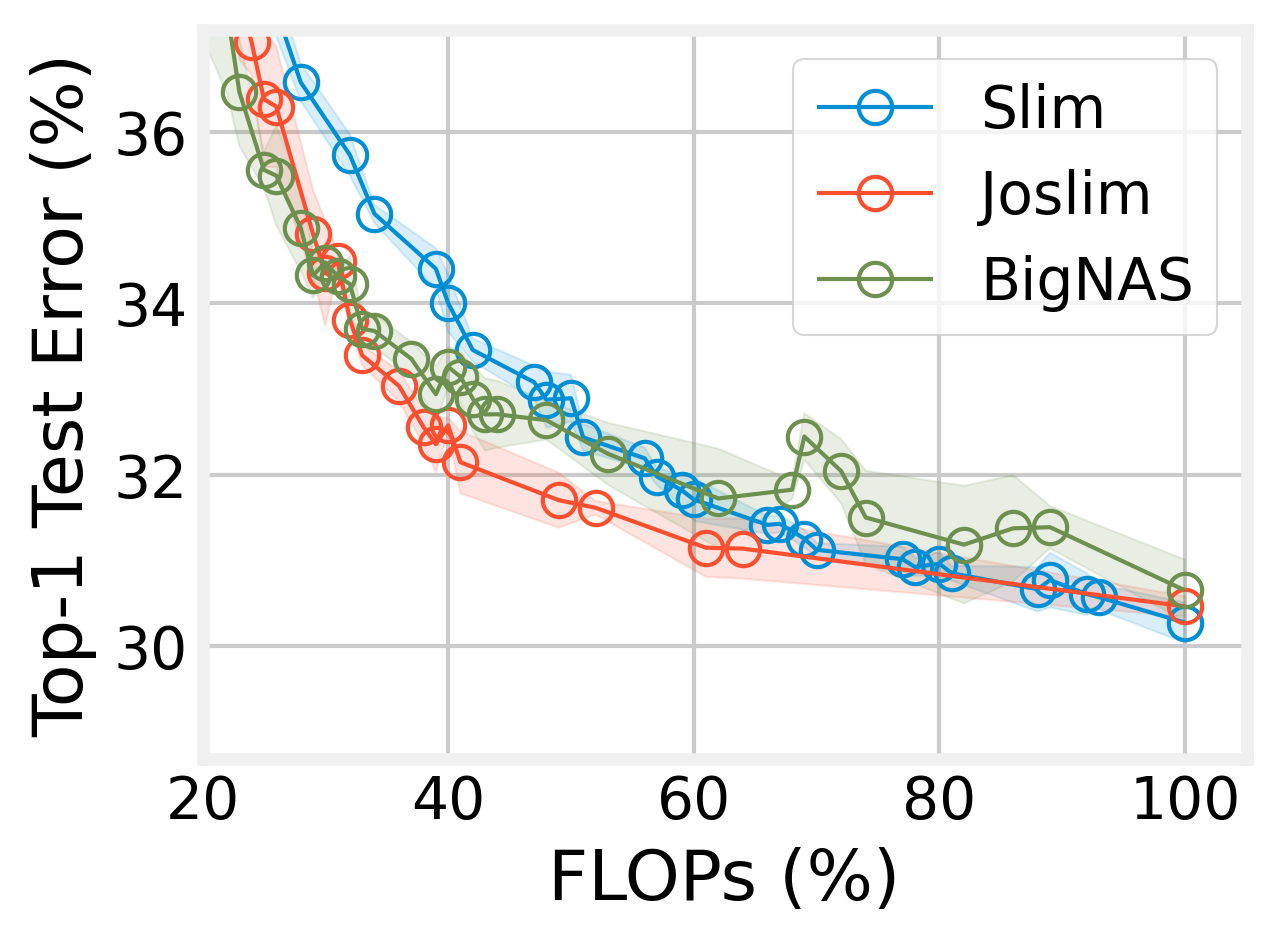}
         \caption{ResNet32 C100}
         \label{resnet32-cifar100}
     \end{subfigure}
     \begin{subfigure}[b]{0.23\textwidth}
         \centering
         \includegraphics[width=\textwidth]{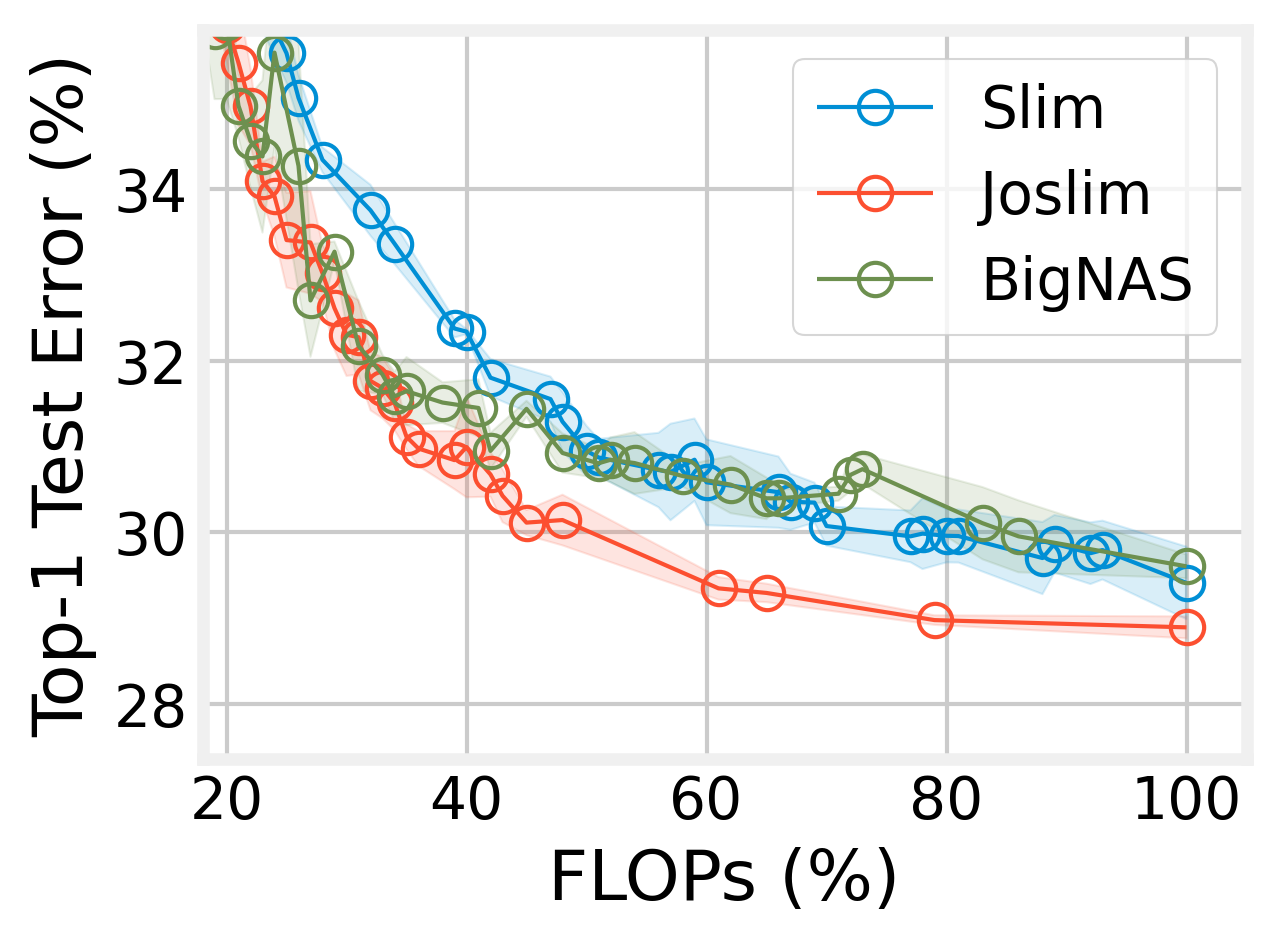}
         \caption{ResNet44 C100}
         \label{resnet44-cifar100}
     \end{subfigure}
     \hfill
     \begin{subfigure}[b]{0.23\textwidth}
         \centering
         \includegraphics[width=\textwidth]{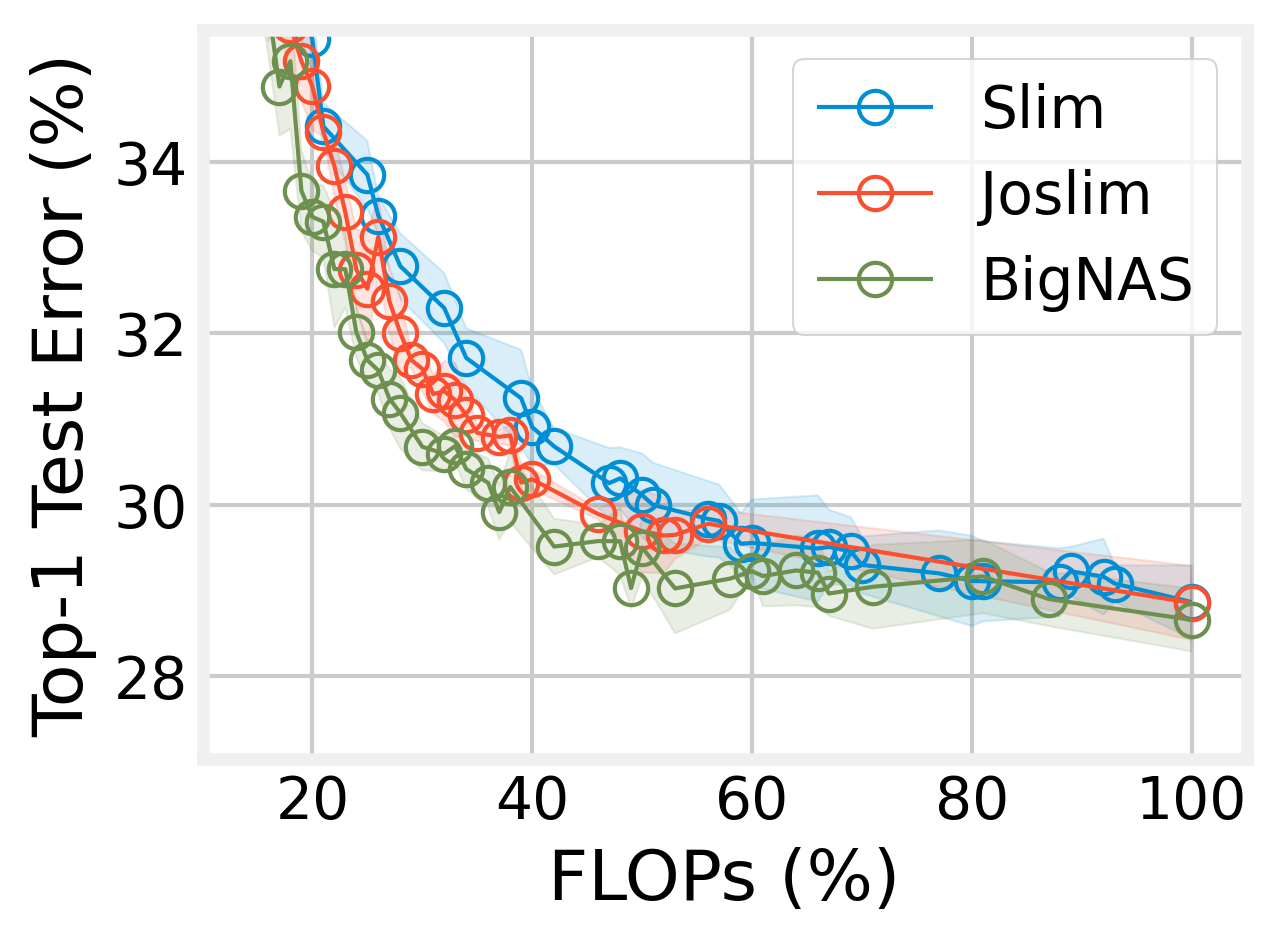}
         \caption{ResNet56 C100}
         \label{resnet56-cifar100}
     \end{subfigure}
     \\
     \begin{subfigure}[b]{0.24\textwidth}
         \centering
         \includegraphics[width=\textwidth]{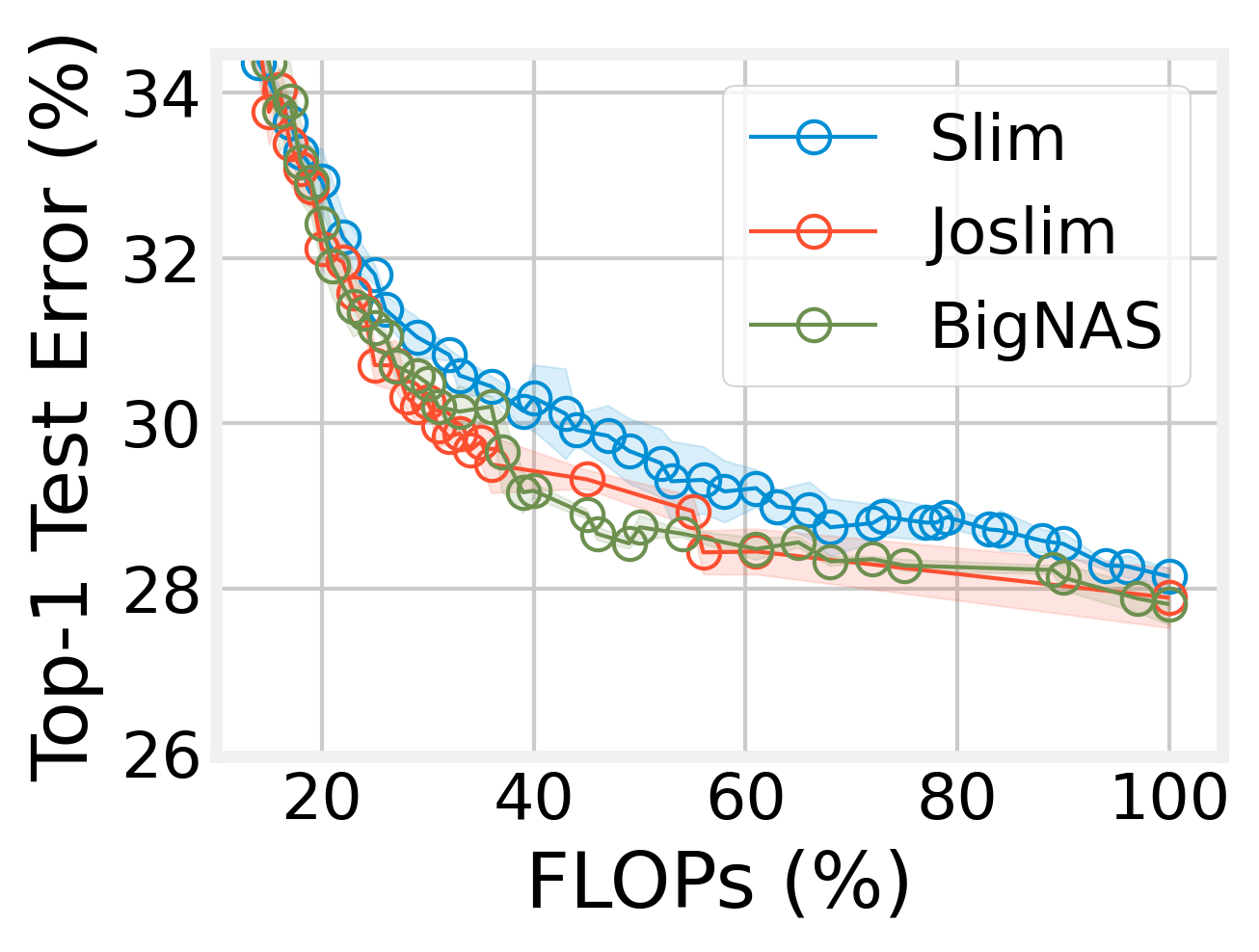}
         \caption{2xResNet20 C100}
         \label{2xresnet20-cifar100}
     \end{subfigure}
     \hfill
     \begin{subfigure}[b]{0.24\textwidth}
         \centering
         \includegraphics[width=\textwidth]{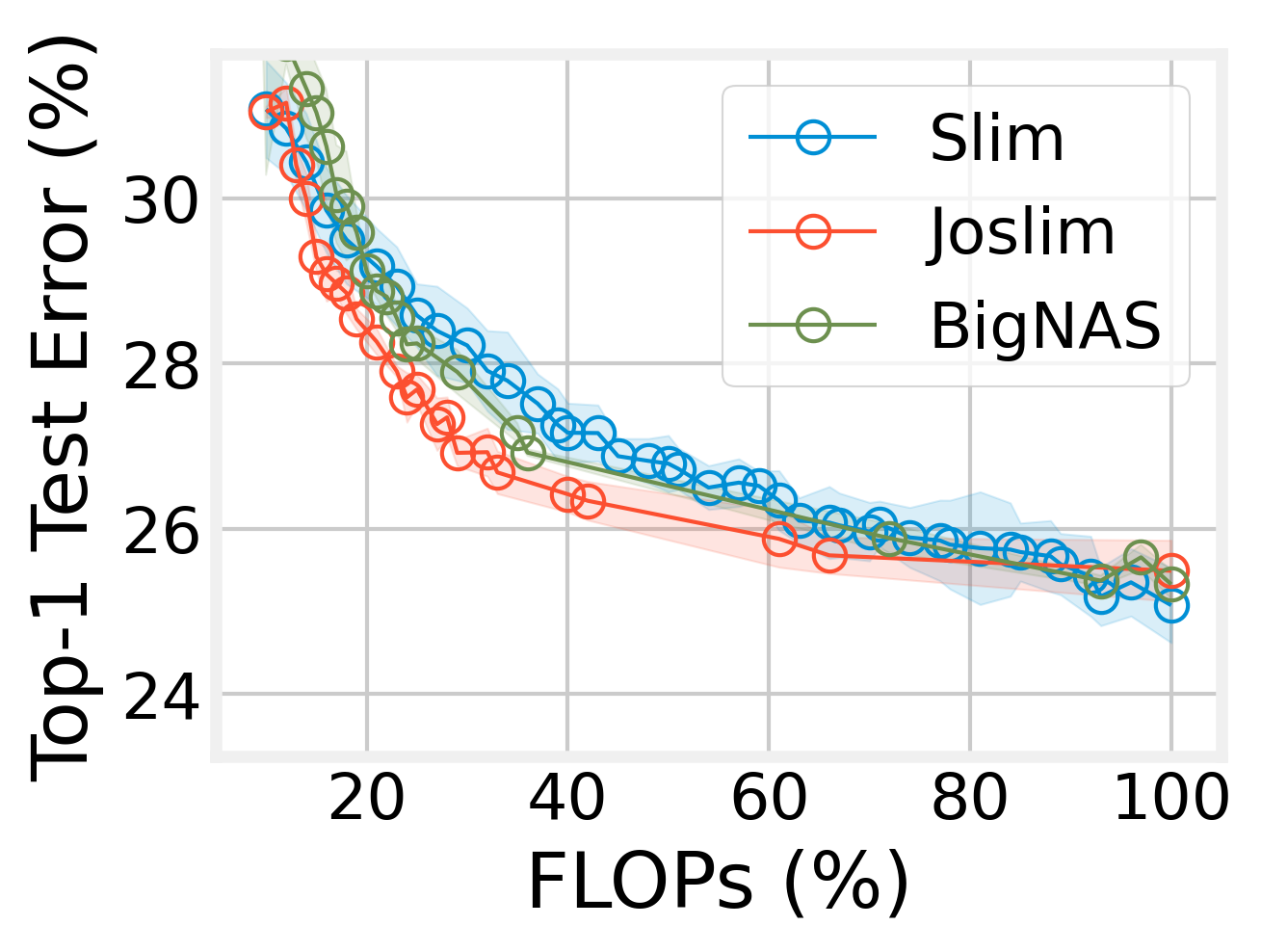}
         \caption{3xResNet20 C100}
         \label{3xresnet20-cifar100}
     \end{subfigure}
     \hfill
     \begin{subfigure}[b]{0.24\textwidth}
         \centering
         \includegraphics[width=\textwidth]{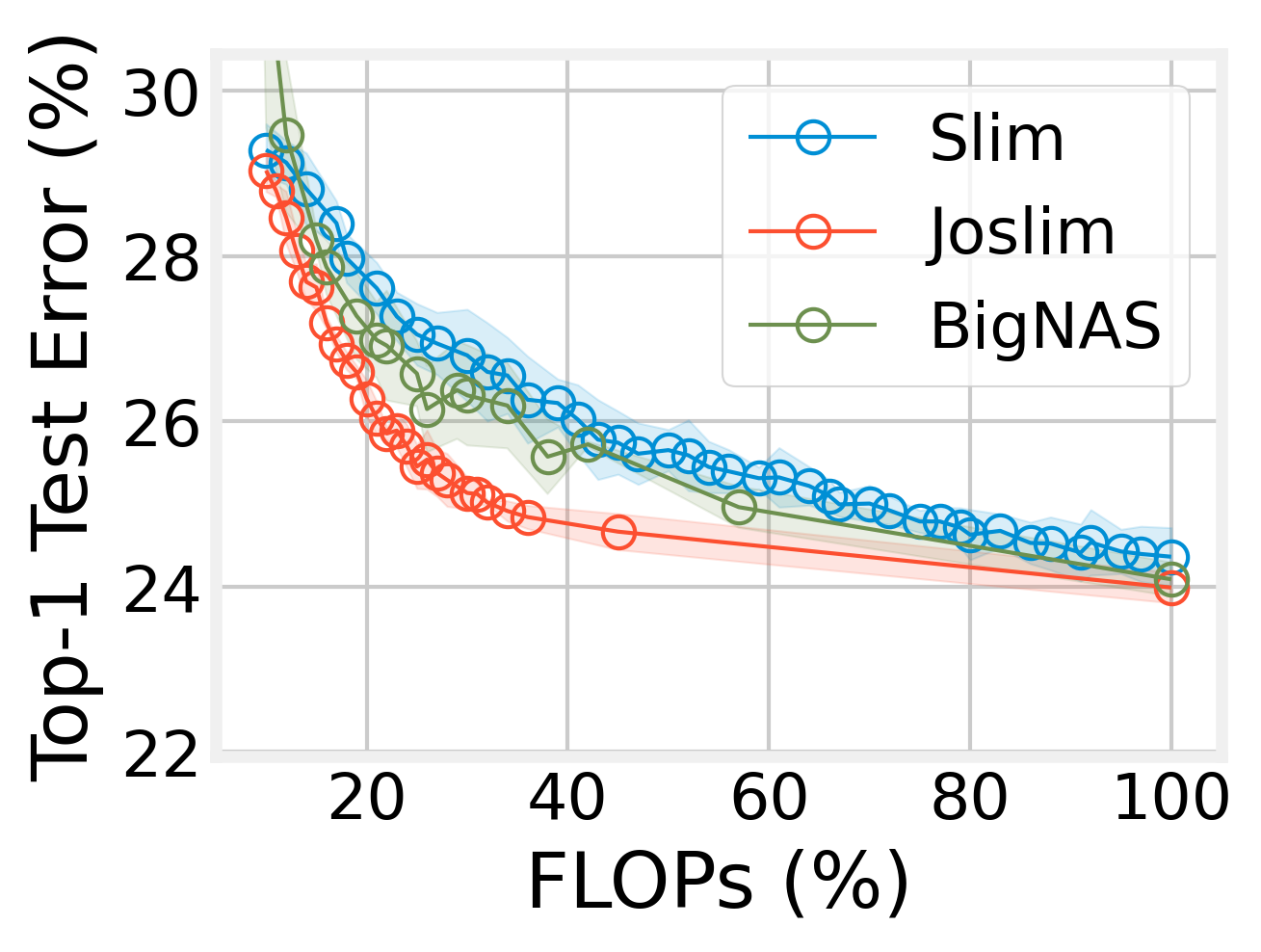}
         \caption{4xResNet20 C100}
         \label{4xresnet20-cifar100}
     \end{subfigure}
     \hfill
     \begin{subfigure}[b]{0.24\textwidth}
         \centering
         \includegraphics[width=\textwidth]{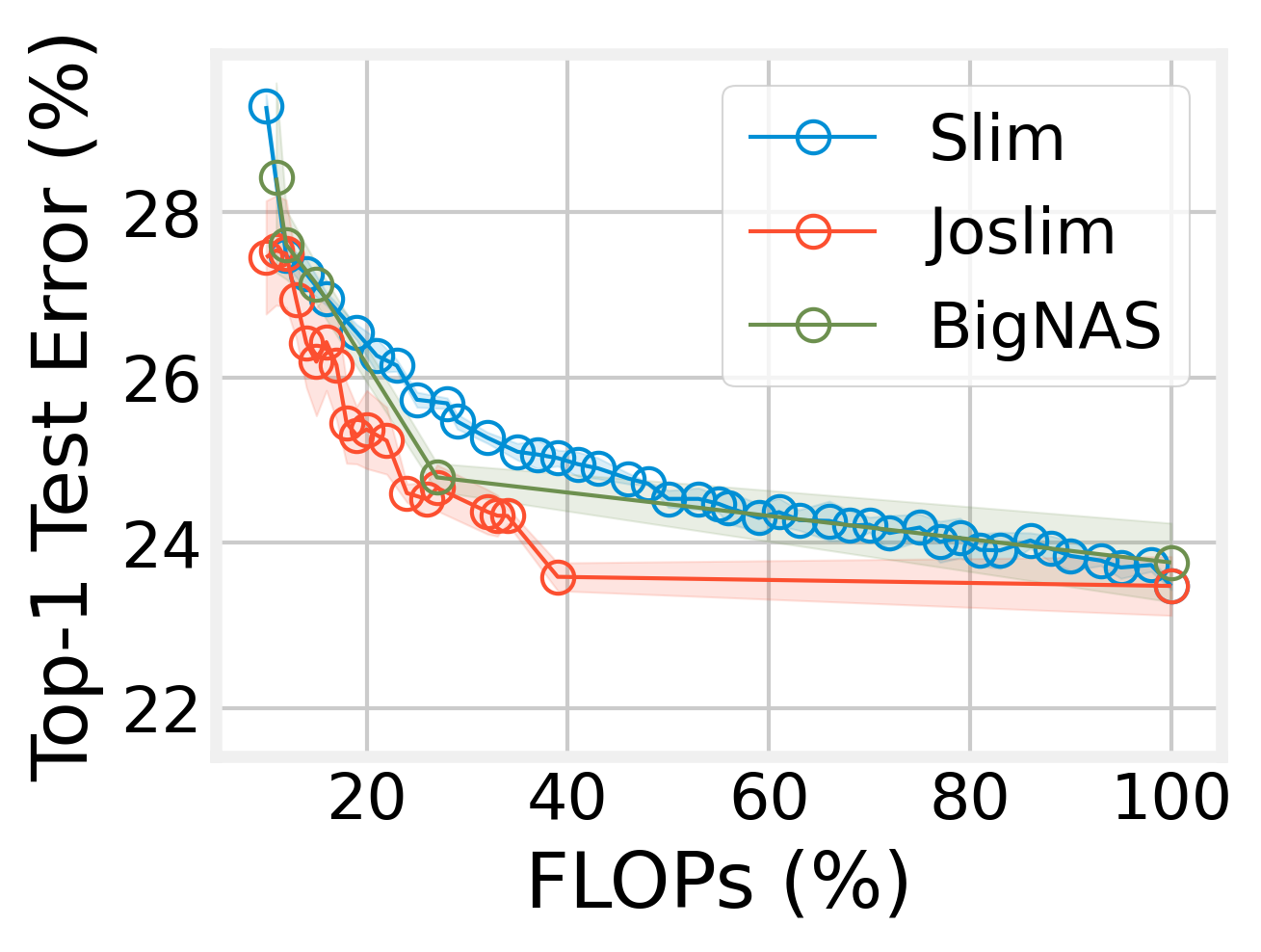}
         \caption{5xResNet20 C100}
         \label{5xresnet20-cifar100}
     \end{subfigure}
    \caption{Comparisons among Slim, BigNAS, and Joslim. C10 and C100 denote CIFAR-10/100. We perform three trials for each method and plot the mean and standard deviation. $n$xResNet20 represents a $n$ times wider ResNet20.}
    \label{fig:resnets-cifar}
\end{figure*}

\begin{figure*}[t!]
     \begin{subfigure}[b]{0.32\textwidth}
         \centering
         \includegraphics[width=\textwidth]{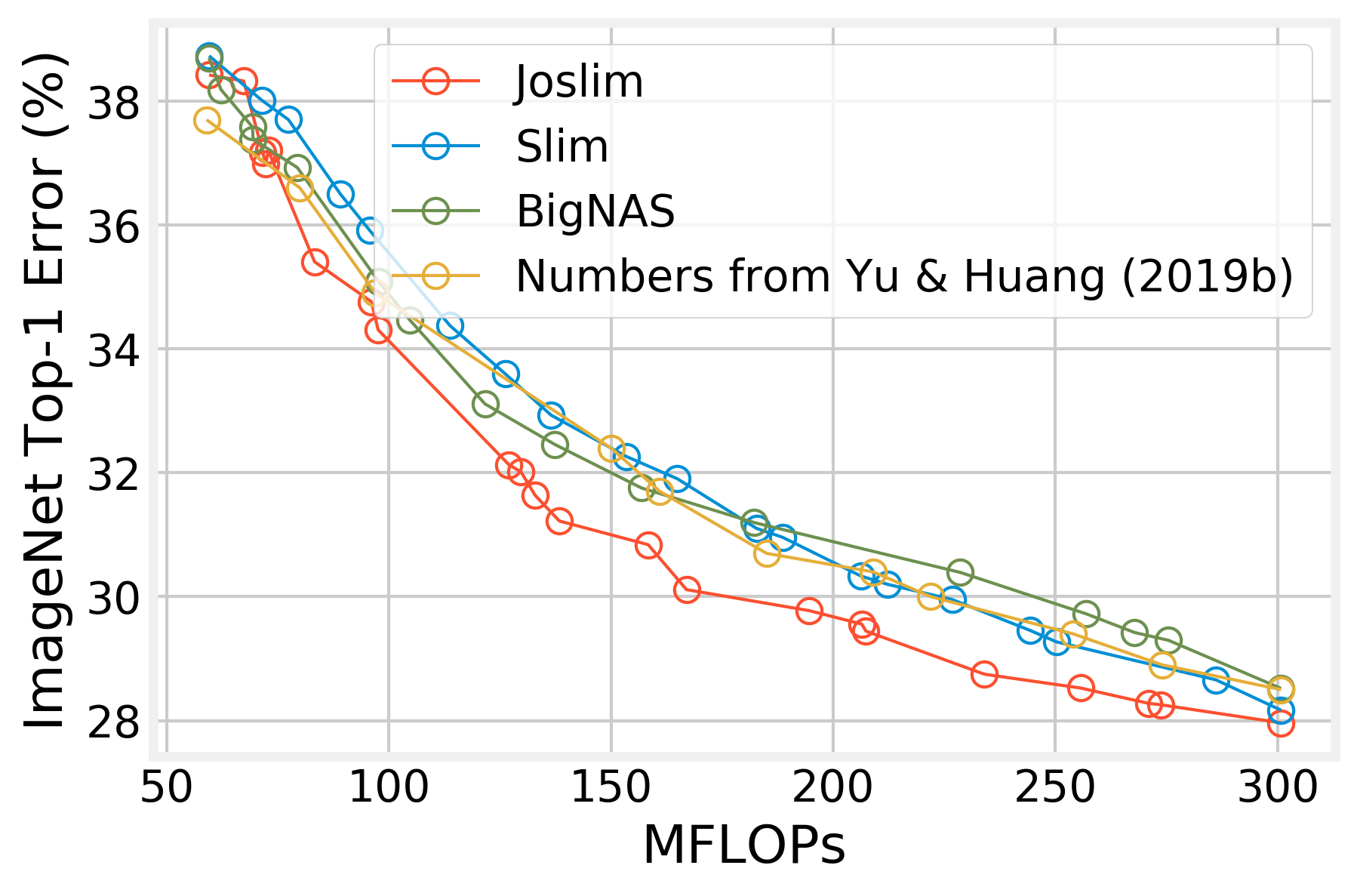}
         \caption{MobileNetV2}
         \label{fig:mbnetv2}
     \end{subfigure}
     \hfill
     \begin{subfigure}[b]{0.32\textwidth}
         \centering
         \includegraphics[width=\textwidth]{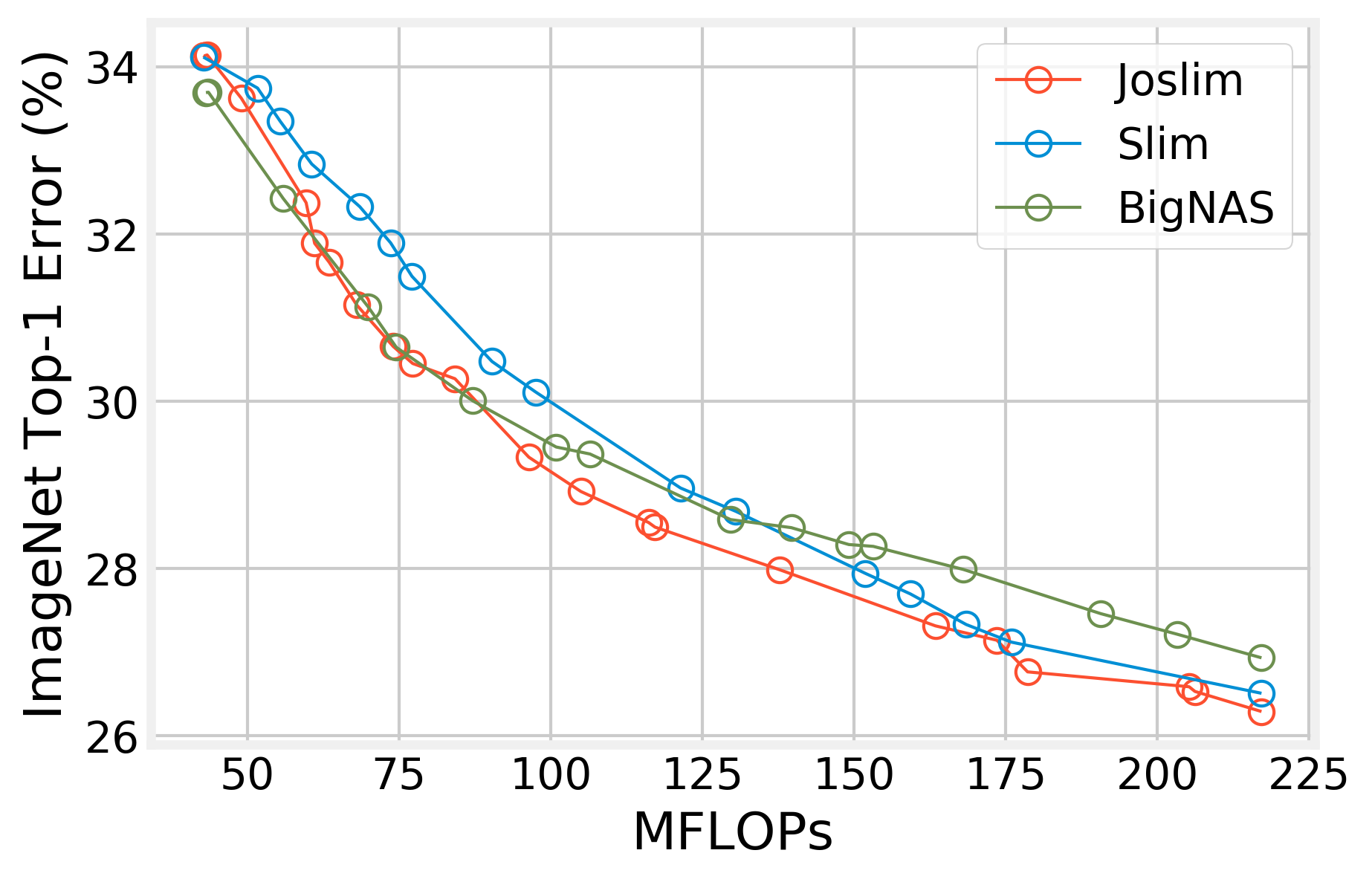}
         \caption{MobileNetV3}
         \label{fig:mbnetv3}
     \end{subfigure}
     \hfill
     \begin{subfigure}[b]{0.32\textwidth}
         \centering
         \includegraphics[width=\textwidth]{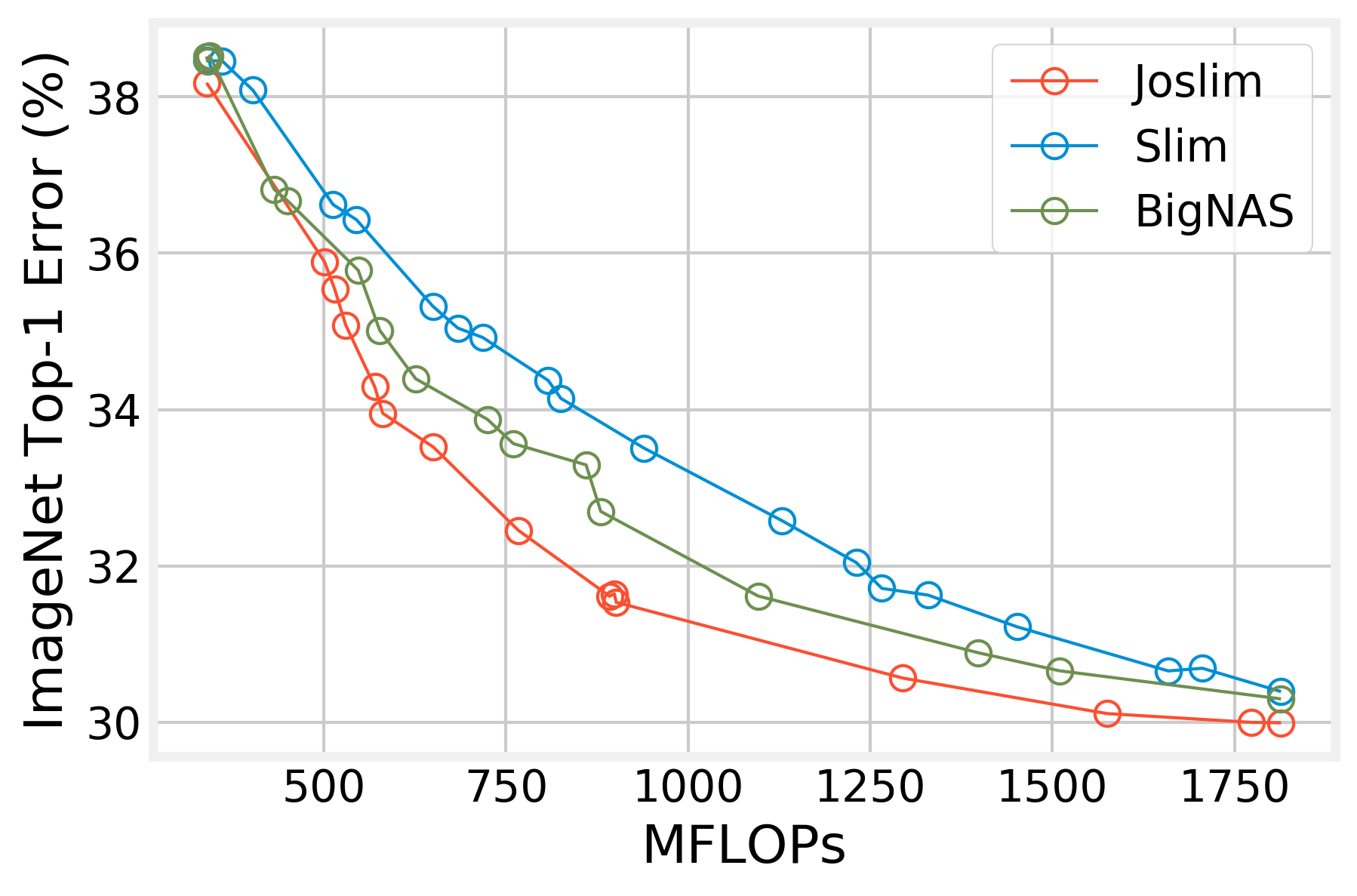}
         \caption{ResNet18}
         \label{fig:res18}
     \end{subfigure}
     
    \caption{Comparisons among Slim, BigNAS, and Joslim on ImageNet.}
    \label{fig:imagenet}
\end{figure*}

We consider three datasets: CIFAR-10, CIFAR-100, and ImageNet. To provide informative comparisons, we verify our implementation for the conventional slimmable training with the reported numbers in~\cite{yu2019universally} using MobileNetV2 on ImageNet. Our results follow closely to the reported numbers as shown in Fig.~\ref{fig:mbnetv2}, which makes our comparisons on other datasets convincing.

We compare to the following baselines:
\begin{itemize}
    \item \textbf{Slim}: the conventional slimmable training method (the universally slimmable networks by~\cite{yu2019universally}). We select 40 architectures uniformly distributed across FLOPs and run a non-dominated sort using training loss and FLOPs.
    \item \textbf{BigNAS}: disjoint optimization that first trains the shared-weights, then uses search methods to find architectures that work well given the trained weights (similar to OFA~\cite{Cai2020Once-for-All:}). To compare fairly with Joslim, we use MOBO-RS for the search. After optimization, we run a non-dominated sort for all the visited architectures $\mathcal{H}$ using training loss and FLOPs.
\end{itemize}

\begin{table*}[t!]
    \centering
    \resizebox{\textwidth}{!}{%
    \begin{tabular}{c|c|c|c||c|c|c|c||c|c|c|c}
        \multicolumn{4}{c||}{MobileNetV2} & \multicolumn{4}{c||}{MobileNetV3} & \multicolumn{4}{c}{ResNet18}\\
        \hline
        MFLOPs & Slim & BigNAS & Joslim & MFLOPs & Slim & BigNAS & Joslim & MFLOPs & Slim & BigNAS & Joslim\\
        \hline
        59 & 61.4 & 61.3 & \textbf{61.5} & 43 & 65.8 & \textbf{66.3} & 65.9 & 339 & 61.5 & 61.5 & \textbf{61.8}\\
        84 & 63.0 & 63.1 & \textbf{64.6} & 74 & 68.1 & 68.1 & \textbf{68.8} & 513 & 63.4 & 64.2 & \textbf{64.5}\\
        102 & 64.7 & \textbf{65.5} & \textbf{65.5} & 85 & 69.1 & \textbf{70.0} & \textbf{70.0} & 650 & 64.7 & 65.6 & \textbf{66.5}\\
        136 & 67.1 & 67.5 & \textbf{68.2} & 118 & 71.0 & \textbf{71.4} & \textbf{71.4} & 718 & 65.1 & 66.1 & \textbf{67.5}\\
        149 & 67.6 & 68.2 & \textbf{69.1} & 135 & 71.5 & 71.5 & \textbf{72.1} & 939 & 66.5 & 67.3 & \textbf{68.5}\\
        169 & 68.2 & 68.8 & \textbf{69.9} & 169 & 72.7 & 72.0 & \textbf{72.8} & 1231 & 68.0 & 68.4 & \textbf{69.4}\\
        212 & 69.7 & 69.6 & \textbf{70.6} & 184 & 73.0 & 72.5 & \textbf{73.2} & 1659 & 69.3 & 69.3 & \textbf{69.9}\\
        300 & 71.8 & 71.5 & \textbf{72.1} & 217 & 73.5 & 73.1 & \textbf{73.7} & 1814 & 69.6 & 69.7 & \textbf{70.0}\\
    \end{tabular}
    }
    \caption{Comparing the top-1 accuracy among Slim, BigNAS, and Joslim on ImageNet. Bold represents the highest accuracy of a given FLOPs.}
    \label{tab:imagenet}
\end{table*}

\begin{figure*}[t]
\centering
\begin{minipage}{.3\textwidth}
    \centering
    \includegraphics[width=\textwidth]{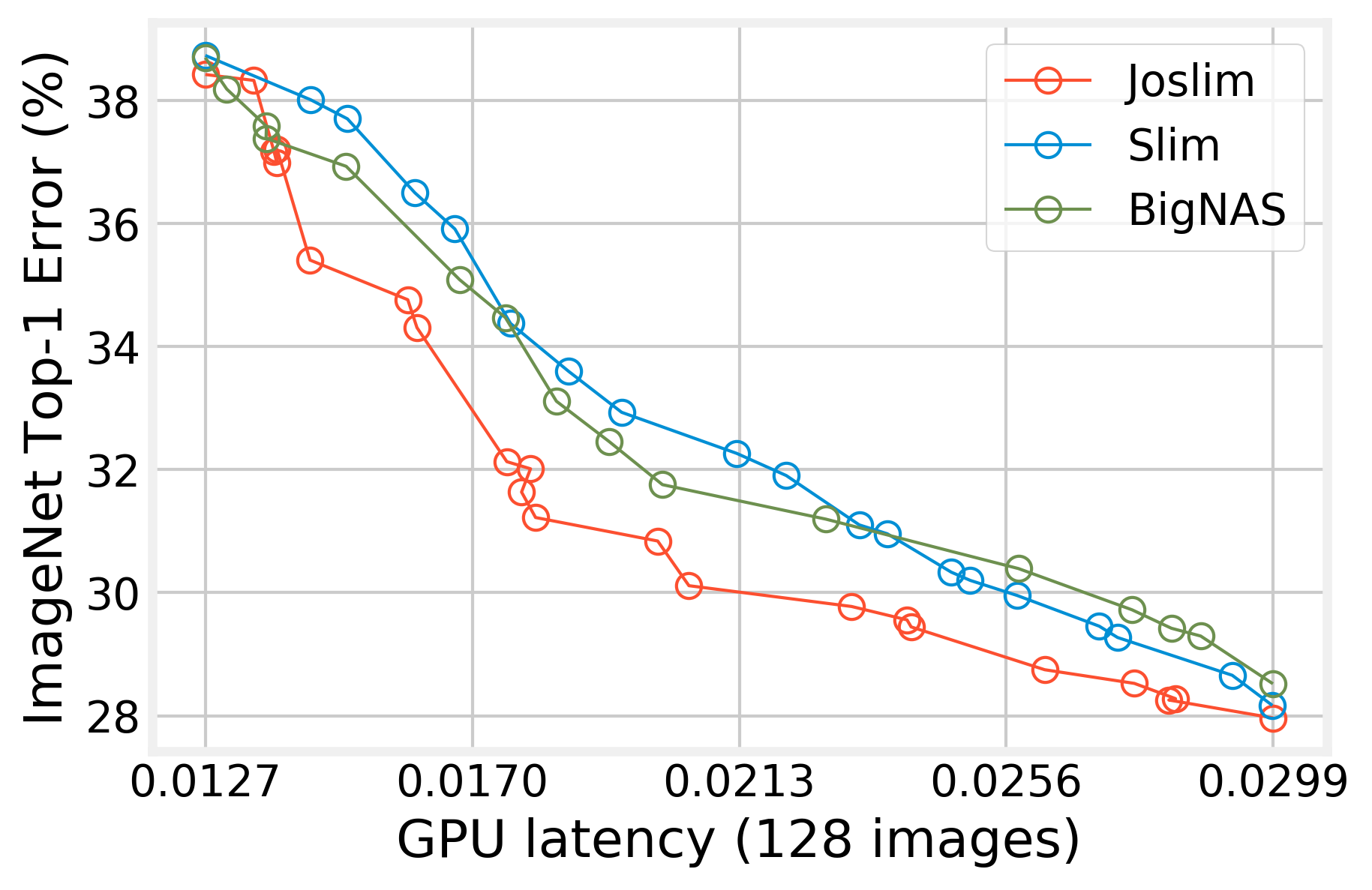}
    \hfill
    \caption{A latency-\emph{vs.}-error view of Fig.~\ref{fig:mbnetv2}.}
    \label{fig:latency}
\end{minipage}%
\hfill
\begin{minipage}{.6\textwidth}
    \centering
    \includegraphics[width=0.45\textwidth]{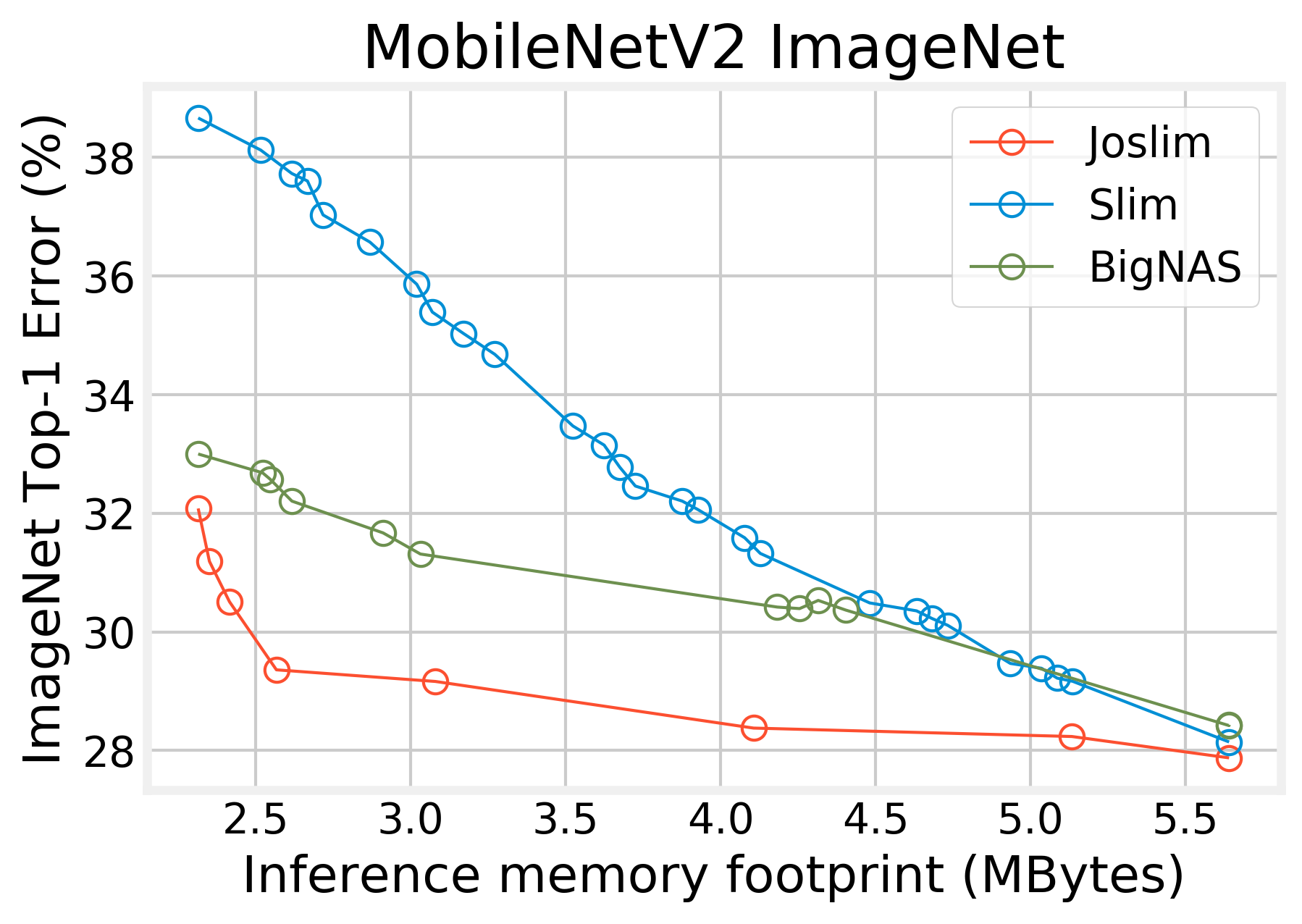}
    \hfill
    \includegraphics[width=0.45\textwidth]{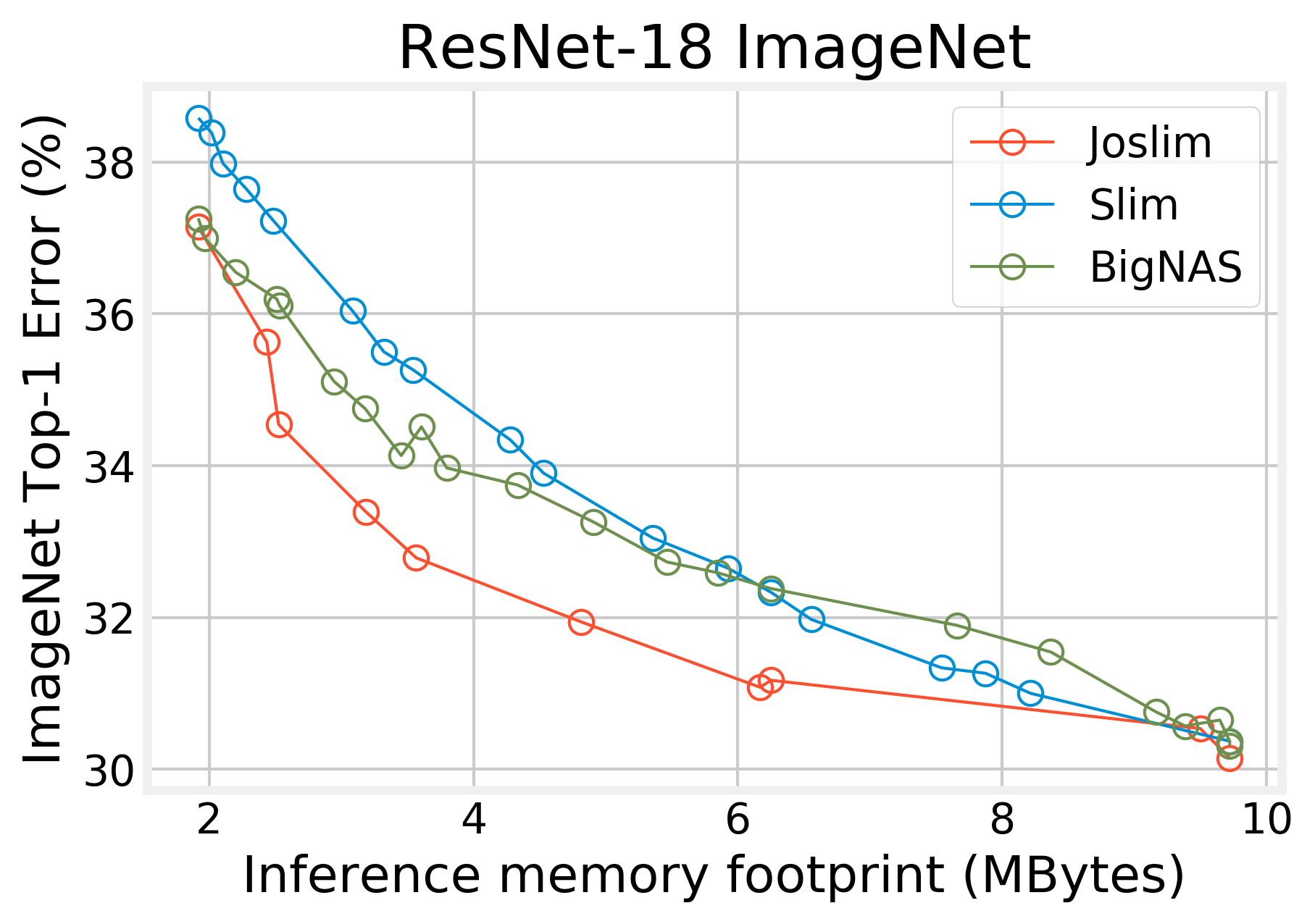}
    \caption{Prediction error \emph{vs.} inference memory footprint for MobileNetV2 and ResNet18 on ImageNet.}
\label{fig:mem}
\end{minipage}
\end{figure*}

\begin{figure*}
     \centering
     \begin{subfigure}[b]{0.23\textwidth}
         \centering
         \includegraphics[width=\textwidth]{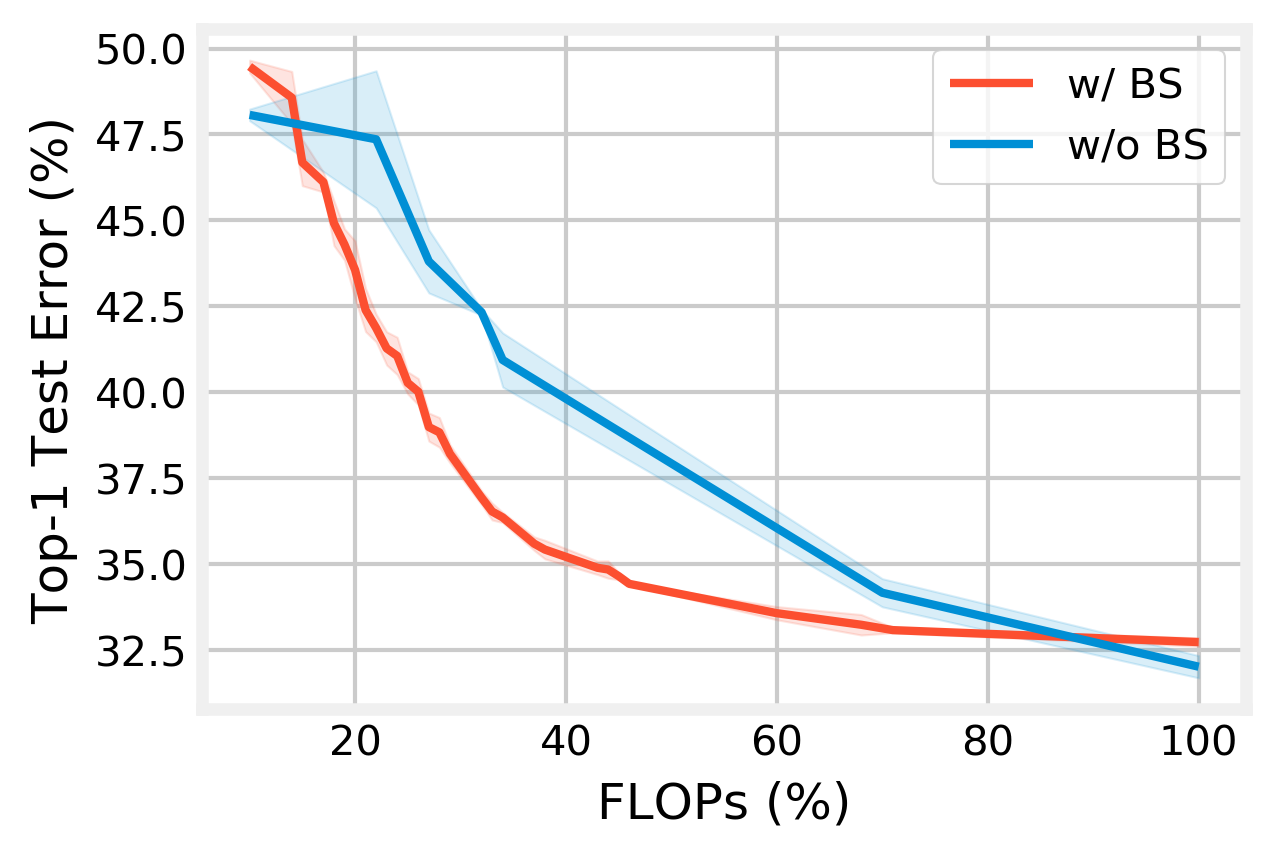}
         \caption{Impact of binary search (BS).}
         \label{pas-perf}
     \end{subfigure}
     \hfill
     \begin{subfigure}[b]{0.23\textwidth}
         \centering
         \includegraphics[width=\textwidth]{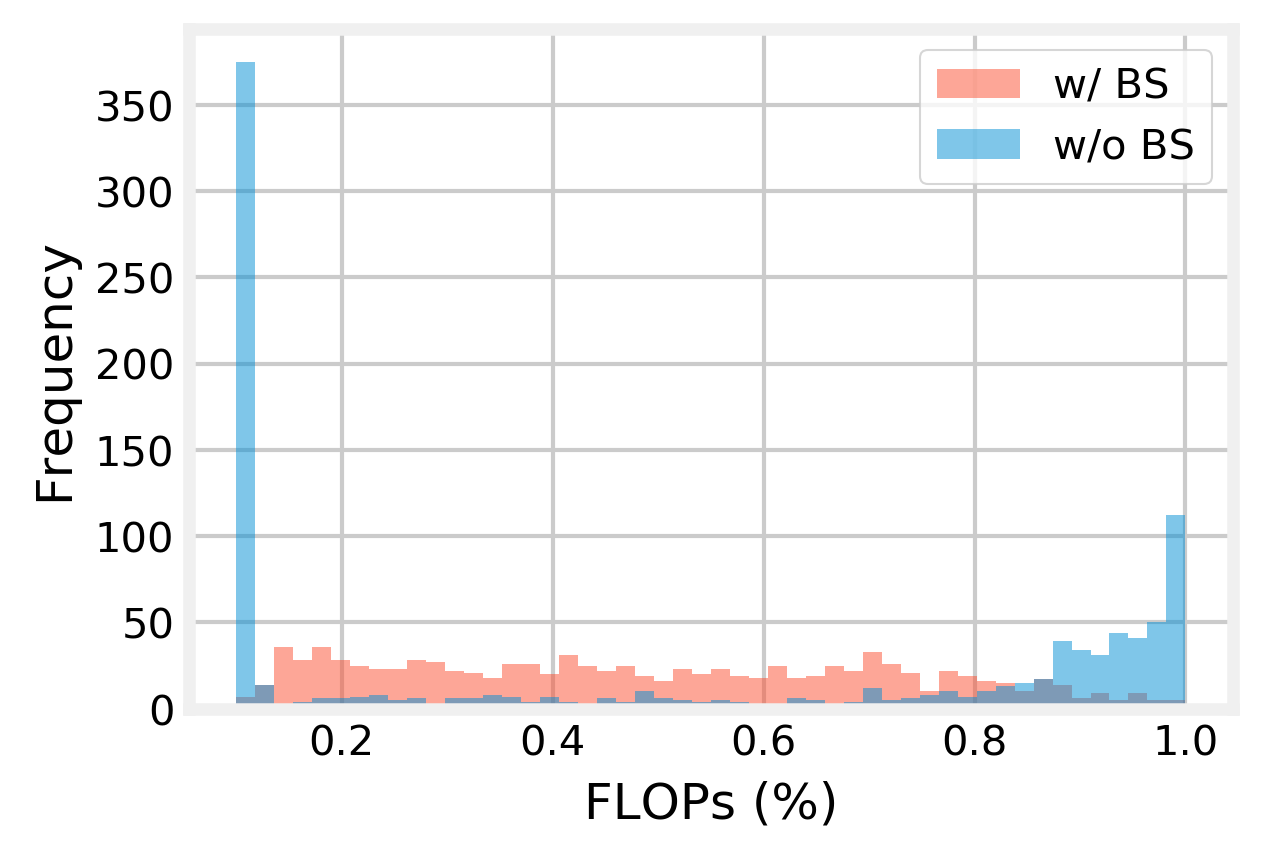}
         \caption{Histogram of FLOPs for $\mathcal{H}$ w/ and w/o BS.}
         \label{pas-hist}
     \end{subfigure}
     \hfill
     \begin{subfigure}[b]{0.23\textwidth}
         \centering
         \includegraphics[width=\textwidth]{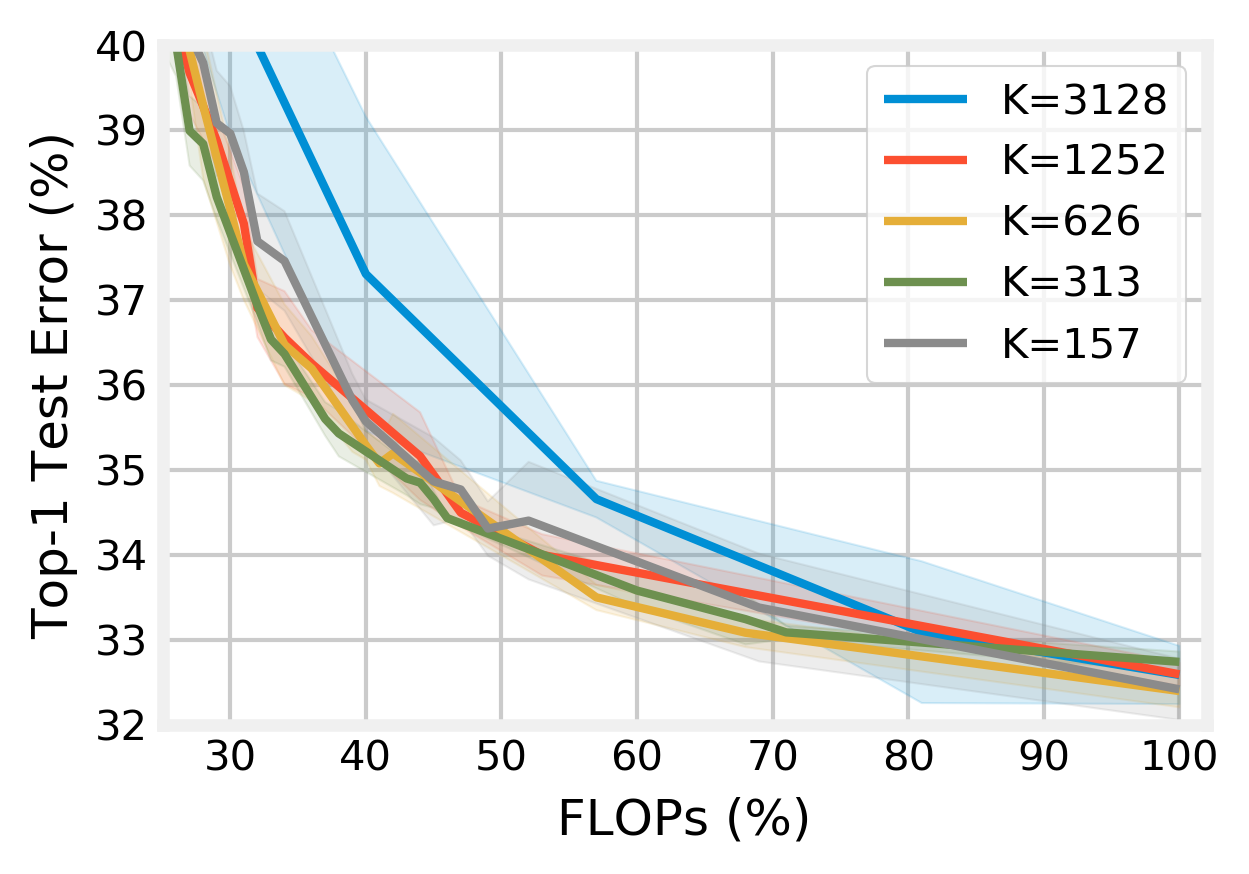}
         \caption{Performance for different $K$.}
         \label{n-ablation}
     \end{subfigure}
     \hfill
     \begin{subfigure}[b]{0.23\textwidth}
         \centering
         \includegraphics[width=\textwidth]{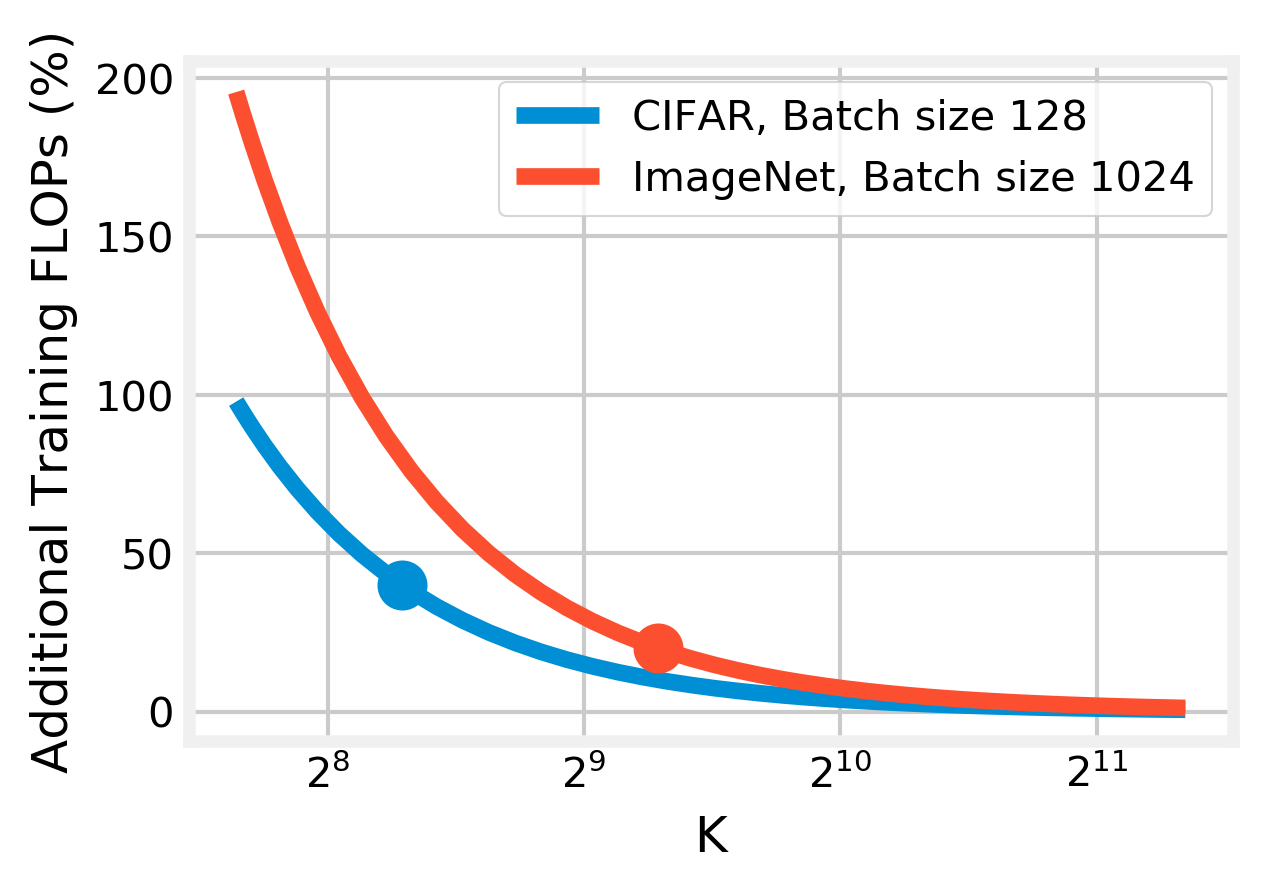}
         \caption{Additional overhead over Slim for different $K$.}
         \label{n-ablation-cost}
     \end{subfigure}
    \caption{Ablation study for the introduced binary search and the number of gradient descent updates per full iteration using ResNet20 and CIFAR-100. Experiments are conducted three times and we plot the mean and standard deviation.}
    \label{fig:ablation}
\end{figure*}

The main results for the CIFAR dataset are summarized in Fig.~\ref{fig:resnets-cifar} with results on ImageNet summarized in Figure~\ref{fig:imagenet} and Table~\ref{tab:imagenet}. Compared to \textit{Slim}, the proposed Joslim has demonstrated much better results across various networks and datasets. This suggests that channel optimization can indeed improve the efficiency of slimmable networks. Compared to \textit{BigNAS}, Joslim is better or comparable across networks and datasets. This suggests that joint widths and weights optimization leads to better overall performance for slimmable nets. From the perspective of training overhead, Joslim introduced minor overhead compared to Slim due to the temporal similarity approximation. More specifically, on ImageNet, Joslim incurs approximately 20\% extra overhead compared to Slim.

Note that the performance among these three methods are similar for the CIFAR-10 dataset. This is plausible since when a network is more over-parameterized, there are many solutions to the optimization problem and it is easier to find solutions with the constraints imposed by weight sharing. In contrast, when the network is relatively less over-parameterized, compromises have to be made due to the constraints imposed by weight sharing. In such scenarios, Joslim outperforms Slim significantly, as it can be seen in CIFAR-100 and ImageNet experiments. We conjecture that this is because Joslim introduces a new optimization variable (width-multipliers), which allows better compromises to be attained. Similarly, from the experiments with ResNets on CIFAR-100 (Fig.~\ref{resnet20-cifar100} to Fig.~\ref{resnet56-cifar100}), we find that shallower models tend to benefit more from joint channel and weight optimization than their deeper counterparts.

As FLOPs may not necessarily reflect latency improvements since FLOP does not capture memory accesses, we in addition plot latency-\emph{vs.}-error for the data in Fig.~\ref{fig:mbnetv2} in Fig.~\ref{fig:latency}. The latency is measured on a single V100 GPU using a batch size of 128. When visualized in latency, Joslim still performs favorably compared to Slim and BigNAS for MobileNetV2 on ImageNet.

Lastly, we consider another objective that is critical for on-device machine learning, \textit{i.e.}, inference memory footprint~\cite{yu2018slimmable}. Inference memory footprint decides whether a model is executable or not on memory-constrained devices. We detailed the memory footprint calculation in Appendix~\ref{sec:mem}. Since Joslim is general, we can replace the FLOPs calculation with memory footprint calculation to optimize for memory-\emph{vs.}-error. As shown in Fig.~\ref{fig:mem}, Joslim significantly outperform other alternatives. Notably, Joslim outperforms Slim by up to 8\% top-1 accuracy for MobileNetV2. Such a drastic improvement comes from the fact that memory footprint depends mostly on the largest layers. As a result, slimming all the layers equally to arrive at networks with smaller memory footprint (as done in Slim) is less than ideal since only one layer contributes to the reduced memory. In addition, when comparing Joslim with BigNAS, we can observe significant improvements as well, \textit{i.e.}, around 2\% top-1 accuracy improvements for MobileNetV2, which demonstrates the effectiveness of joint width and weights optimization.

\subsection{Ablation studies}\label{sec:ablation}
In this subsection, we ablate the hyperparameters that are specific to Joslim to understand their impact. We use ResNet20 and CIFAR-100 for the ablation with the results summarized in Fig.~\ref{fig:ablation}.

\subsubsection{Binary search} Without binary search, one can also consider sampling the scalarization weighting $\lambda$ uniformly from $[0,1]$, which does not require any binary search and is easy to implement. However, the issue with this sampling strategy is that uniform sampling $\lambda$ does not necessarily imply uniform sampling in the objective space, \textit{e.g.}, FLOPs. As shown in Fig.~\ref{pas-perf} and Fig.~\ref{pas-hist}, sampling directly in the $\lambda$ space results in non-uniform FLOPs and worse performance compared to binary search.

\subsubsection{Number of gradient descent steps} In the approximation, the number of architectures ($|\mathcal{H}|$) is affected by the number of gradient descent updates $K$. In previous experiments for CIFAR, we have $K=313$, which results in $|\mathcal{H}|=1000$. Here, we ablate $K$ to $156,626,1252,3128$ such that $|\mathcal{H}|=2000,500,250,100$, respectively. Given a fixed training epoch and batch size, Joslim produces a better approximation for equation~\ref{eq:weight_sample} but a worse approximation for equation~\ref{eq:arch_sample} with larger $K$. The former is because of the local approximation while the latter is because there are overall fewer iterations put into Bayesian optimization due to temporal sharing. As shown in Fig.~\ref{n-ablation}, we observe worse results with higher $K$. On the other hand, the improvement introduced by lower $K$ saturates quickly. The overhead of Joslim as a function of $K$ compared to Slim is shown in Fig.~\ref{n-ablation-cost} where the dots are the employed $K$.
 
\section{Conclusion}
In this work, we are interested in optimizing both the architectural components and shared-weights of slimmable neural networks. To achieve this goal, we propose a general framework that optimizes slimmable nets by minimizing the area under the trade-off curve between cross entropy and FLOPs (or memory footprint) with alternating minimization. We further show that the proposed framework subsumes existing methods as special cases and provides flexibility for devising better algorithms. To this end, we propose Joslim, an algorithm that jointly optimizes the weights and widths of slimmable nets, which empirically outperforms existing alternatives that either neglect width optimization or conduct widths and weights optimization independently. We extensively verify the effectiveness of Joslim over existing techniques on three datasets (\textit{i.e.}, CIFAR10, CIFAR100, and ImageNet) with two families of network architectures (\textit{i.e.}, ResNets and MobileNets) using two types of objectives (\textit{i.e.}, FLOPs and memory footprint). Our results highlight the importance and superiority in results of jointly optimizing the channel counts for different layers and the weights for slimmable networks.

\section*{Acknowledgement}
This research was supported in part by NSF CCF Grant No. 1815899, NSF CSR Grant No. 1815780, and NSF ACI Grant No. 1445606 at the Pittsburgh Supercomputing Center (PSC).

\bibliography{main}
\bibliographystyle{splncs04}

\clearpage
\appendix

\section{Width parameterization}\label{app:wm}
For ResNets with CIFAR, $\bm{a}$ has six dimensions and is denoted by $\bm{a}_{1:6}\in[0.316, 1]$, \textit{i.e.}, one parameter for each stage and one for each residual connected layers in three stages. More specifically, the network is divided into three stages according to the output resolution, and as a result, there are three stages for all the ResNets designed for CIFAR. For example, in ResNet20, there are 7, 6, and 6 layers for each of the stages, respectively. Also, the layers that are added together via residual connection have to share the same width-multiplier, which results in one width-multiplier per stage for the layers that are connected via residual connections.

For MobileNetV2, $\bm{a}_{1:25}\in[0.42, 1]$, and therefore there is one dimension for each independent convolutional layer. Note that while there are in total 52 convolutional layers in MobileNetV2, not all of them can be altered independently. More specifically, for layers that are added together via residual connection, their widths should be identical. Similarly, the depth-wise convolutional layer should have the same width as its preceding point-wise convolutional layers. The same logic applies to MobileNetV3, which has 47 convolutional layers (excluding squeeze-and-excitation layers) and $\bm{a}_{1:22}\in[0.42, 1]$. In MobileNetV3, there are squeeze-and-excitation (SE) layers and we do not alter the width for the expansion layer in the SE layer. The output width of the SE layer is set to be the same as that of the convolutional layer where the SE layer is applied to. Note that there is no concept of expansion ratio for the inverted residual block in MobileNets in our width optimization. More specifically, the convolutional layer that acts upon expansion ratio is in itself just a convolutional layer with tunable width. Also, we do not quantize the width to be multiples of $8$ as adopted in the previous work~\cite{sandler2018mobilenetv2,yu2019universally}. Due to these reasons, our $0.42\times$ MobileNetV2 has 59 MFLOPs, which has the same FLOPs as the $0.35\times$ MobileNetV2 in~\cite{yu2019universally,sandler2018mobilenetv2}.

\section{Training hyperparameters}\label{app:train}
We use PyTorch as our deep learning framework and we build MOBO-TS on BoTorch~\cite{balandat2020botorch}, which works seamlessly with PyTorch. More specifically, for the covariance function of Gaussian Processes, we use the commonly adopted Mat{\'e}rn Kernel without changing the default hyperparameters provided in BoTorch. Similarly, we use $beta=0.1$ for the Upper Confidence Bound acquisition function. To perform the optimization of line 6 in Algorithm~\ref{alg:pas}, we make use of the API ``\textit{optimize\_acqf}" provided in BoTorch. As a reference, with a single 1080Ti GPU, one can train a Joslim-ResNet20 on CIFAR-100 with around 3 hours. On the other hand, with 8 V100 GPUs on a single machine, one can train a Joslim-ResNet18 on ImageNet with 19 hours.

\subsubsection{CIFAR} The training hyperparameters for the independent models are 0.1 initial learning rate, 200 training epochs, 0.0005 weight decay, 128 batch size, SGD with nesterov momentum, and cosine learning rate decay. The accuracy on the validation set is reported using the model at the final epoch. For slimmable training, we keep the same exact hyperparameters but train $2\times$ longer compared to independent models, \textit{i.e.}, 400 epochs. 

\subsubsection{ImageNet} Our training hyperparameters follow that of~\cite{yu2019universally}. Specifically, we use initial learning rate of 0.5 with 5 epochs linear warmup (from 0 to 0.5), linear learning rate decay (from 0.5 to 0), 250 epochs, $4e^{-5}$ weight decay, 0.1 label smoothing, and we use SGD with 0.9 nesterov momentum. We use a batch size of 1024. For data augmentation, we use the ``RandomResizedCrop'' and ``RandomHorizontalFlip'' APIs in PyTorch. For MobileNetV2 we follow~\cite{yu2019universally} and use random scale between 0.25 to 1. For MobileNetV3, we use the default scale parameters, \textit{i.e.}, from 0.08 to 1. The input resolution we use is 224. Besides scaling and horizontal flip, we follow~\cite{yu2019universally} and use color and lighting jitters data augmentataion with parameter of 0.4 for brightness, contrast, and saturation; and 0.1 for lighting. These augmentations can be found in the official repository of~\cite{yu2019universally}\footnote{\url{https://github.com/JiahuiYu/slimmable_networks/blob/master/train.py\#L43}}. The hyperparamters for training ResNet18 is identical to MobileNetV2 except that we train it for 100 epochs only. The training for ImageNet is done using 8 NVIDIA V100 GPUs.

\section{Width differences}\label{app:arch}
\begin{figure*}[ht]
     \centering
     \begin{subfigure}[b]{0.48\textwidth}
         \centering
         \includegraphics[width=0.9\textwidth]{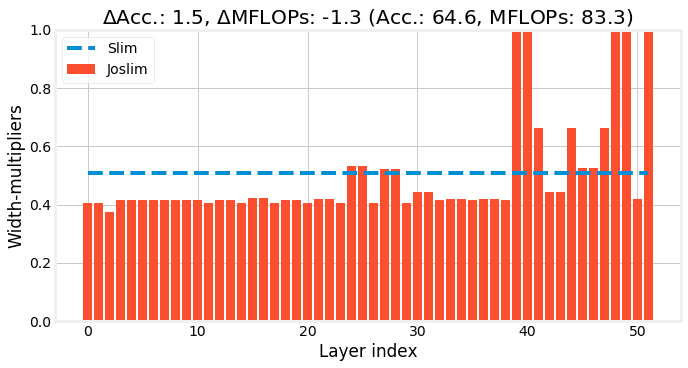}
     \end{subfigure}
     \hfill
     \begin{subfigure}[b]{0.48\textwidth}
         \centering
         \includegraphics[width=0.9\textwidth]{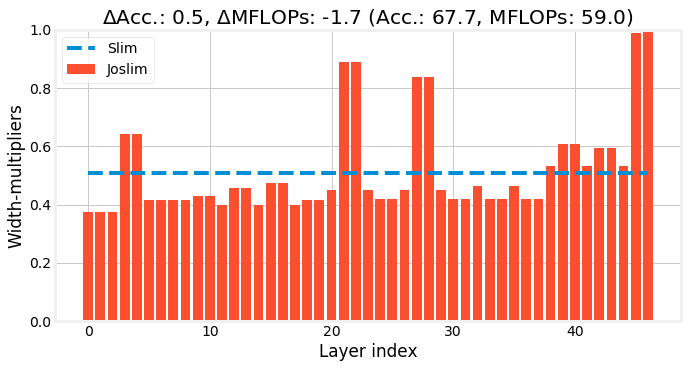}
     \end{subfigure}
     \\
     \begin{subfigure}[b]{0.48\textwidth}
         \centering
         \includegraphics[width=0.9\textwidth]{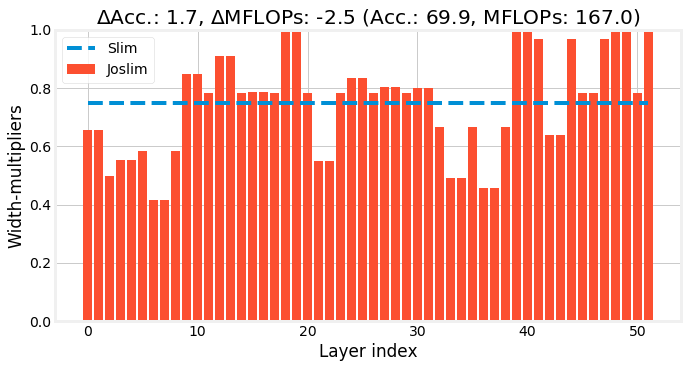}
     \end{subfigure}
     \hfill
     \begin{subfigure}[b]{0.48\textwidth}
         \centering
         \includegraphics[width=0.9\textwidth]{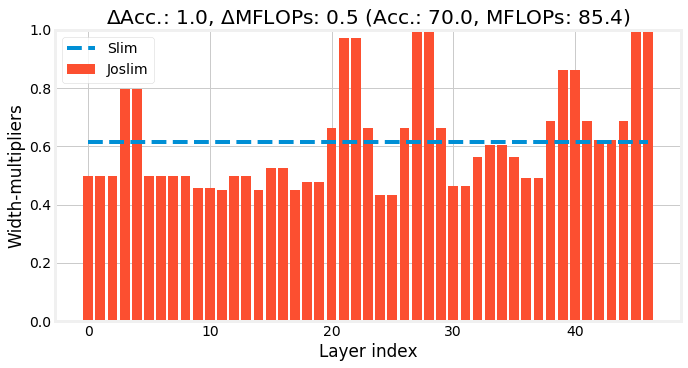}
     \end{subfigure}
     \\
     \begin{subfigure}[b]{0.48\textwidth}
         \centering
         \includegraphics[width=0.9\textwidth]{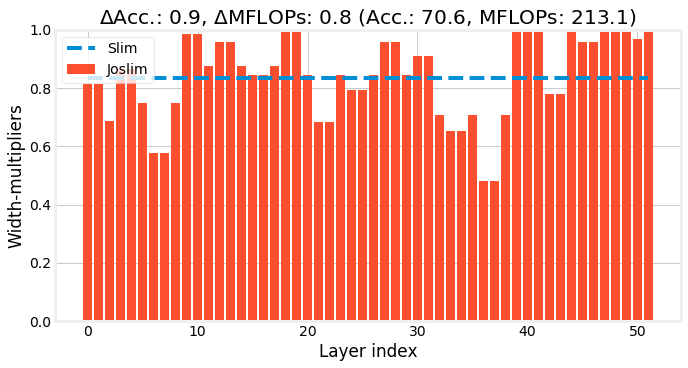}
         \caption{MobileNetV2 ImageNet}
     \end{subfigure}
     \hfill
     \begin{subfigure}[b]{0.48\textwidth}
         \centering
         \includegraphics[width=0.9\textwidth]{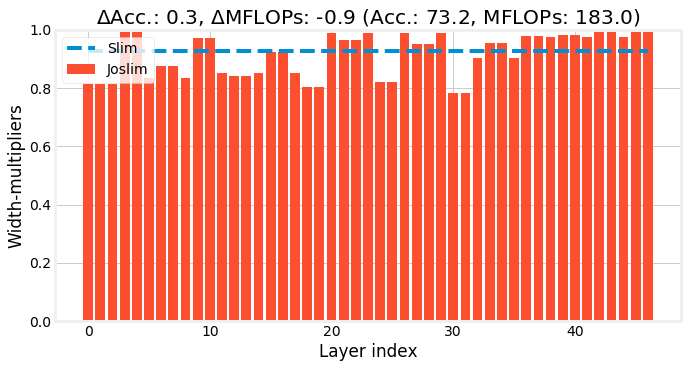}
         \caption{MobileNetV3 ImageNet}
     \end{subfigure}
     
    \caption{Comparing the width-multipliers between Joslim and Slim. The title for each plot denotes the relative differences (Joslim - Slim) and the numbers in the parenthesis are for Joslim.}
    \label{fig:width}
\end{figure*}

In Fig.~\ref{fig:width}, we visualize the widths learned by Joslim and contrast them with Slim for MobileNetV2 and MobileNetV3. Note that both Joslim and Slim are slimmable networks with shared weights and from the top row to the bottom row represent three points on the trade-off curve for Fig.~\ref{fig:mbnetv2} and Fig.~\ref{fig:mbnetv3}.

\section{Theoretical analysis for temporal approximation}\label{app:bonud}
The intuition behind the proposed approximation in Section~\ref{sec:joslim} is the similarity for $\bm{\theta}$ across alternating minimization. In an extreme case, if we hold $\bm{\theta}$ constant throughout the training procedure, the approximation is equivalent to the original multi-objective BO. With that said, $\bm{\theta}$ changes gradually throughout training. To proceed with further theoretical understanding, we assume the loss $L_{\mathcal{S}}(\bm{\theta})$ is L-Lipschitz. More formally,
\begin{align}
    \begin{split}
        L_{\mathcal{S}}(\bm{\theta}^t) - L_{\mathcal{S}}(\bm{\theta}^{t+1}) \leq L \|\bm{\theta}^t - \bm{\theta}^{t+1}\|_1,\forall \bm{\theta}^t,\bm{\theta}^{t+1}.
    \end{split}
\end{align}
Now, consider using stochastic gradient descent to update the weights $\bm{\theta}$, \emph{i.e.}, $\bm{\theta}^{t+1}=\bm{\theta}^t-\eta^t \bm{g}^t$ where $\bm{g}^t$ is the gradient of loss with respect to the weights and $\eta^t$ is the learning rate at iteration $t$. Since $L_{\mathcal{S}}$ is L-Lipschitz, we have $\|\bm{g}\|_1\leq L$. Assuming using an exponential decaying learning rate with a factor $\gamma < 1$, we can further upper bound the functional differences across $K$ iterations of gradient descents as follows:
\begin{align}
    \begin{split}
         L_{\mathcal{S}}(\bm{\theta}^t) - L_{\mathcal{S}}(\bm{\theta}^{t+n}) \leq \sum_{i=t}^{t+K}\eta^i \|g^i\| \leq K \eta^t L.
    \end{split}
\end{align}

Aligning with our intuition, the analysis reveals that larger $K$ implies poorer approximation for Bayesian optimization to share history. In multi-objective Bayesian optimization~\cite{PariaKP19}, the hyperparameter is searched over stationary objectives. In our case, due to temporal approximation, our cross entropy changes over time and the change is upper-bounded by $K \eta^t L$. As a result, we can plug such an upper bound in the regret bound analysis of Bayesian optimization~\cite{PariaKP19} to understand how $K$, $\eta$, and $\gamma$ affect the optimality of Bayesian optimization. Specifically, we upper bound $L_{\mathcal{S}}(\bm{\theta}^t)$ with $L_{\mathcal{S}}(\bm{\theta}^{t+K}) + K \eta^t L$ and use it in Lemma 2 and Lemma 3 from \cite{PariaKP19} in Appendix B.1. With such a technique, a regret bound will have the following overhead in addition to the original regret bound in equation (14) of \cite{PariaKP19}:
\begin{align}
    \begin{split}
        \frac{2\eta^0}{1-\gamma}K L\mathbb{E}[L_{\lambda}]K',
    \end{split}
\end{align}
where we have utilized the geometric progression of the exponential learning rate decay and $L$, $\mathbb{E}[L_{\lambda}]$, and $K'$ are the notations used by~\cite{PariaKP19}. In other words, without a decaying learning rate, the overhead can be unbounded. This analysis reveals that larger initial learning rate $\eta^0$ and $K$ results in a worse regret bound.

\section{Inference memory footprint calculation}\label{sec:mem}
To demonstrate the generality of proposed Joslim, we in addition consider optimizing for the trade-off curve between prediction error and inference memory footprint. The inference memory footprint is a critical factor when it comes to deploying deep CNNs onto resource-constrained devices such as mobile phones or micro-controllers as motivated in the original slimmable neural network paper~\cite{yu2018slimmable}. We use a single image per batch to calculate the memory footprint. Specifically the inference memory footprint is characterized as follows:
\begin{align}
    \begin{split}
        FM_{in}^l & = W_{in}^l \times H_{in}^l \times C_{in}^l\\
        FM_{out}^l & = W_{out}^l \times H_{out}^l \times C_{out}^l\\
        \text{Weights}^l & = K_w^l \times K_h^l \times C_{in}^l \times C_{out}^l / G^l\\
        \text{Skip}^l & = W_{out}^l \times H_{out}^l \times C_{skip}^l\\
        MEM & = \max_l \left( FM_{in}^l+FM_{out}^l+\text{Weights}^l+\text{Skip}^l \right),
    \end{split}
\end{align}
where $FM_{in}^l$ and $FM_{out}^l$ denote the input and output feature map sizes of layer $l$, $\text{Weights}^l$ denotes the size of the weights in layer $l$, and $\text{Skip}^l$ denotes the memory requirement of storing the feature maps from skip connections. $W$ and $H$ represent the width and height of the feature map. $K_w$ and $K_h$ denote the kernel size. Lastly, $C_{in}$, $C_{out}$ and $G$ denote the input channel, output channel, and the number of groups for convolutional layer $l$.

\subsection{Comparisons with AutoSlim}\label{app:autoslim}
\begin{figure}
     \centering
     \begin{subfigure}[b]{0.4\textwidth}
         \centering
         \includegraphics[width=\textwidth]{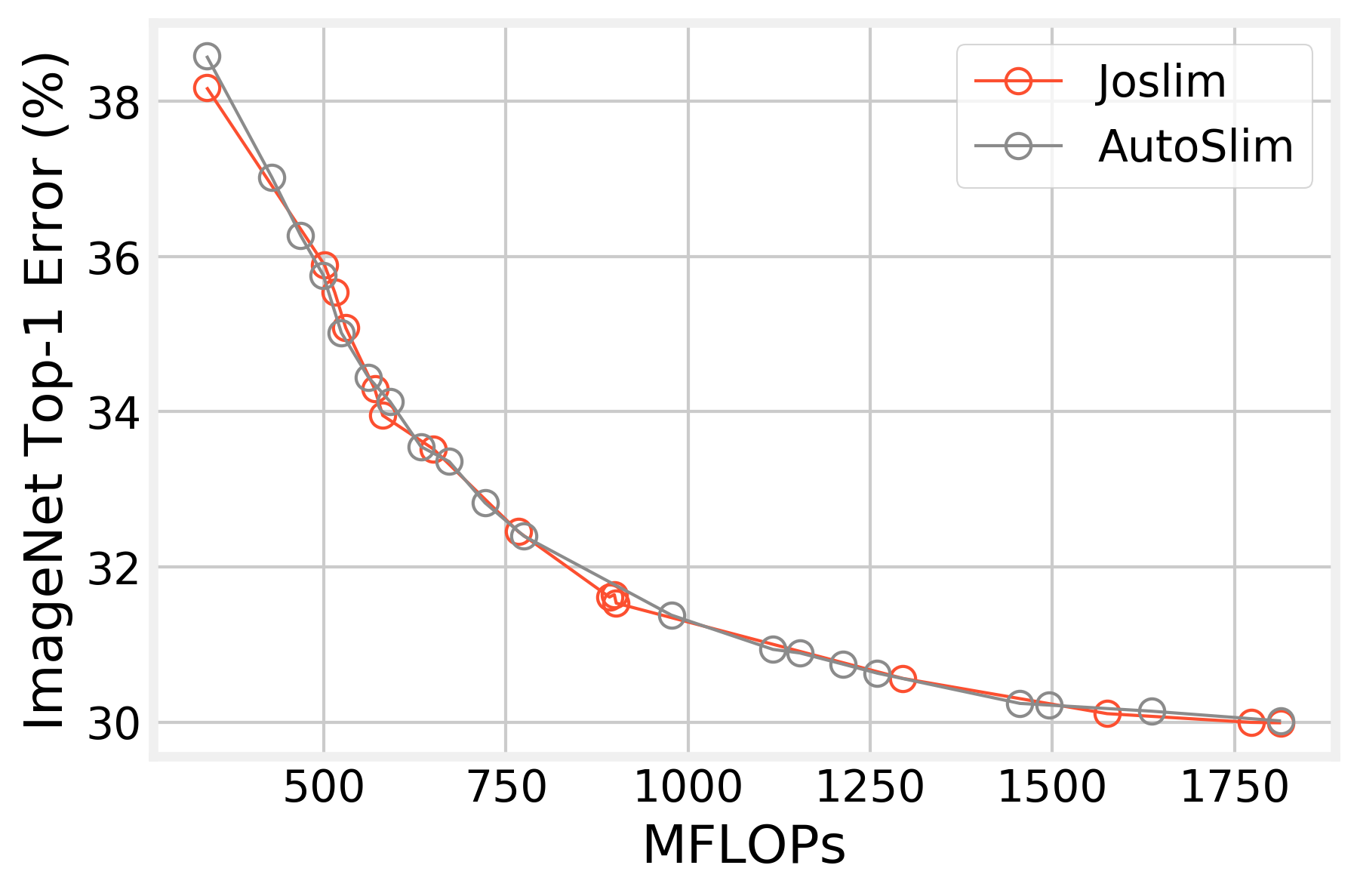}
         \caption{Error \textit{vs.} FLOPs}
     \end{subfigure}
     \hfill
     \begin{subfigure}[b]{0.4\textwidth}
         \centering
         \includegraphics[width=\textwidth]{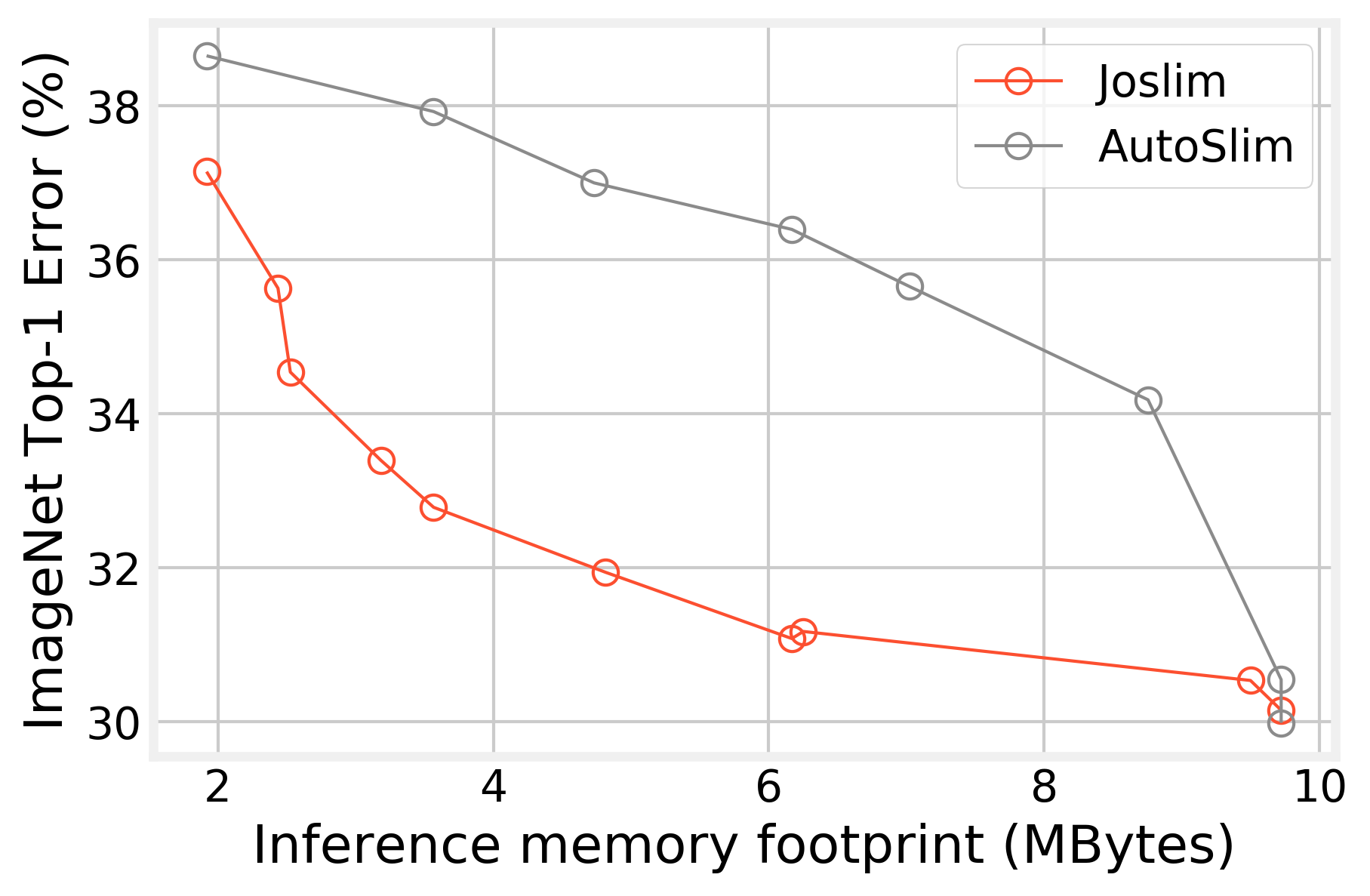}
         \caption{Error \textit{vs.} Memory}
     \end{subfigure}
    \caption{Comparing Joslim and AutoSlim on ResNet18. Since ResNet18 has similar FLOPs across different layers, greedy pruning used by AutoSlim perform comparably to Joslim. However, Joslim outperforms AutoSlim when it comes to optimizing for memory consumption since the greedy pruning procedure adopted by AutoSlim is not multi-objective.}
    \label{fig:autoslim}
\end{figure}

AutoSlim~\cite{yu2019autoslim} is a NAS method proposed to do channel search for \textit{standalone models}. While it also provides non-uniform widths for different layers, it is not a method derived to solve for equation~\ref{eq:prob}. Specifically, AutoSlim first perform weight-sharing training (equation~\ref{eq:weight}) for a short amount of period, i.e., 10\% to 20\% of the full training epochs. Then, AutoSlim conducts greedy pruning to greedily remove a fixed amount of channels from the layer that affects the loss the least. Such a greedy procedure naturally results in a sequence of models of different computational requirements. Crucially, this greedy pruning procedure is not solving equation~\ref{eq:arch_set} since the computational requirement (FLOPs or memory footprint) does not affect the ranking among filters to be pruned. Nonetheless, we can adopt AutoSlim to obtain a sequence of models and train them via weight-sharing to form a slimmable network. We compared with AutoSlim using ResNet18 on ImageNet with both FLOPs and memory footprint. As shown in Fig.~\ref{fig:autoslim}, Joslim performs similarly with AutoSlim when it comes to FLOPs, and this is because ResNet18 has balanced FLOPs for all the layers. On the other hand, when it comes to memory footprint, AutoSlim performs much worse compared to Joslim. In hindsight, this result is not surprising as Joslim is designed to solve equation~\ref{eq:prob} while AutoSlim is not.

\end{document}